\let\accentvec\vec
\documentclass[runningheads]{llncs}

\let\vec\accentvec
\usepackage{graphicx}
\usepackage[T1]{fontenc}
\usepackage{amsmath} 
\usepackage{amssymb}
\usepackage{color}
\usepackage{multirow}
\usepackage{subfigure}

\usepackage{cvpr_eso}
\usepackage[width=122mm,left=12mm,paperwidth=146mm,height=193mm,top=12mm,paperheight=217mm]{geometry}
\newcommand{\tabincell}[2]{\begin{tabular}{@{}#1@{}}#2\end{tabular}}

\begin{document}

\title{STH: Spatio-Temporal Hybrid Convolution \\
for Efficient Action Recognition} 


\titlerunning{STH: Spatio-Temporal Hybrid Convolution for Efficient Action Recognition}
%
\author{Xu Li\inst{1}\thanks{Work done while Xu Li was a Research Intern with Tencent AI Lab.}, Jingwen Wang\inst{2}$^{\dag}$, Lin Ma\inst{2}$^{\dag}$, Kaihao Zhang\inst{3}, Fengzong Lian\inst{2}, \\
Zhanhui Kang\inst{2} and Jinjun Wang\inst{1}$^{\dag}$
}
\authorrunning{X. Li, J. Wang, L. Ma, K. Zhang, F. Lian, Z. Kang and J. Wang.}
%
\institute{Xi'an Jiaotong University \and Tencent AI Lab \and Australian National University \\
\email{\{im.leexu,jaywongjaywong,forest.linma\}@gmail.com,kaihao.zhang@anu.edu.au}
\email{\{faxonlian,kegokang\}@tencent.com,jinjun@mail.xjtu.edu.cn}
}

\renewcommand{\thefootnote}{\dag}
\footnotetext{Corresponding authors.}

\maketitle

\begin{abstract}
Effective and Efficient spatio-temporal modeling is essential for action recognition. Existing methods suffer from the trade-off between model performance and model complexity. In this paper, we present a novel Spatio-Temporal Hybrid Convolution Network (denoted as ``STH'') which simultaneously encodes spatial and temporal video information with a small parameter cost. Different from existing works that sequentially or parallelly extract spatial and temporal information with different convolutional layers, we divide the input channels into multiple groups and interleave the spatial and temporal operations in one convolutional layer, which deeply incorporates spatial and temporal clues. Such a design enables efficient spatio-temporal modeling and maintains a small model scale. STH-Conv is a general building block, which can be plugged into existing 2D CNN architectures such as ResNet and MobileNet by replacing the conventional 2D-Conv blocks (2D convolutions). STH network achieves competitive or even better performance than its competitors on benchmark datasets such as Something-Something (V1 \& V2), Jester, and HMDB-51. Moreover, STH enjoys performance superiority over 3D CNNs while maintaining an even smaller parameter cost than 2D CNNs.

\end{abstract}

\section{Introduction}

The research community has witnessed significant progresses in action recognition in the past decade. Up to date the accuracy of action recognition has been greatly improved \cite{lgd,i3d}. Although these models enjoy good performances, their model sizes and computation complexity could be unfavorable, especially for embedded deployment (e.g., mobile phones). As videos are increasingly popular with the advances of digital cameras and Internet, it requires certain models to simultaneously maintain high accuracy, high computational efficiency and acceptable storage cost. In this paper, we aim to develop a more powerful spatio-temporal model for action recognition with low complexity.

\begin{figure}[t]
\centering
\includegraphics[width=0.65\textwidth]{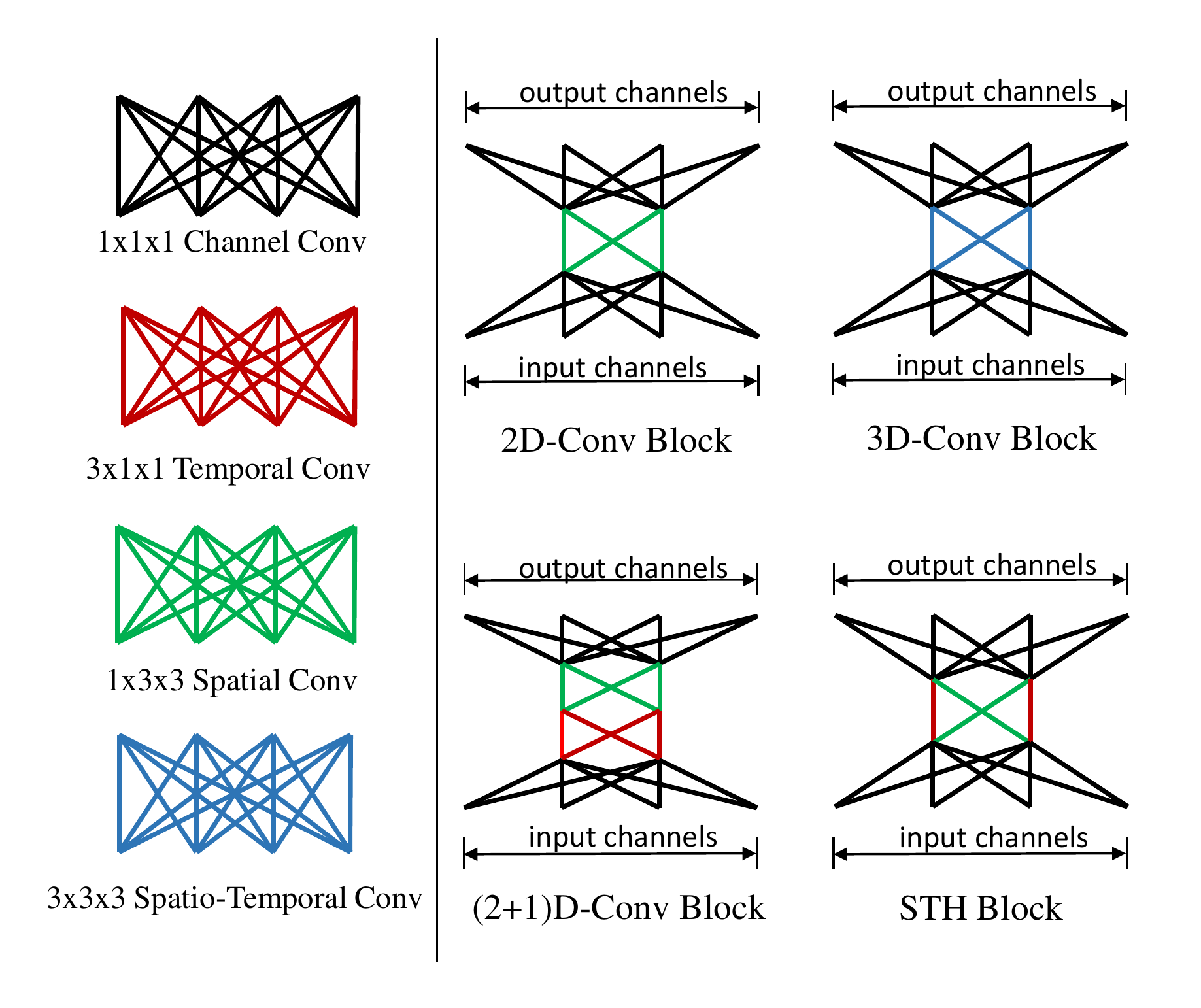}
\vspace{-15pt}
\caption{Comparison of our proposed STH block with existing convolutional blocks. 2D-Conv adopts spatial kernels to only capture appearance features. 3D-Conv adopts three-dimensional kernels to encode spatio-temporal information. (2+1)D-Conv sequentially stacks spatial and temporal kernels. In contrast, our proposed STH-Conv constructs spatio-temporal hybrid kernels by interleaving basic spatial and temporal kernels along the input channel \textbf{in one convolutional layer}.}
\vspace{-10pt}
\label{figure_introduction}
\end{figure}

Generally speaking, spatial information focus on static appearance features such as actors and objects from a video, while temporal information can be regarded as an indicator for recognizing motion and action. State-of-the-art approaches leverage both spatial and temporal information to enhance the performance of action recognition \cite{lgd,tsm,i3d,r2+1d,slowfast,dynamonet}. Temporal modeling is of key importance for recognizing actions. For example, the actions ``\texttt{\small{take up}}'' and ``\texttt{\small{put down}}'' can only be differentiated with the aid of temporal information. The conventional 2D CNNs is good at modeling spatial information but cannot capture temporal clues. One solution is the so-called ``two-stream'' methods \cite{lifeifei,2stream,2streamfusion,tsn,trn}, which apply one extra 2D CNN to learn motion features based on pre-extracted optical flow. However, extracting optical flow is notorious for its inefficiency and storage cost. Temporal Segment Network leverages the idea of temporal structure modeling but it can only perceive long-range motion change \cite{tsn}. The follow-up solution is 3D CNNs \cite{c3d,i3d}, which is capable of directly learning spatio-temporal features from $N$ consecutive frames. While augmenting 2D-Conv to 3D-Conv is simple, it is not computationally efficient due to the high-dimensional 3D kernels \cite{3dcnn}. Figure \ref{figure_introduction} graphically illustrates the difference between different convolutions.

As illustrated above, 2D CNNs lack temporal modeling capability while 3D CNNs are computationally intensive. Some researchers attempt to reach a compromise between performance and complexity by mixing 2D and 3D CNNs (Mixed 2D\&3D CNNs) \cite{eco,mixed2d3d} or decomposing 3D convolutions into 2D spatial convolutions followed by 1D temporal convolutions ((2+1)D CNNs) \cite{s3d,p3d,r2+1d}. However, among these approaches, the mixed 2D\&3D CNNs reduce computational complexity at the expense of restricting temporal modeling capability, while the spatio-temporal separation convolutions learn spatial and temporal information using two sequentially-stacked layers. We argue that it is better to construct deeper coupling of the two operations (see Table \ref{table_somethingv1} for performance comparison).

To this end, we propose a novel convolution called Spatio-Temporal Hybrid Convolution (denoted as ``STH''), which effectively encodes spatial and temporal clues for action recognition. As shown in Figure \ref{figure_introduction}, conventional convolutional blocks, including 2D-Conv, 3D-Conv, and (2+1)D-Conv, directly augment 1D/2D/3D kernels along the channel dimension. For example, the shape of an augmented convolutional kernel of a 2D-Conv layer is $C_i \times H \times W$ (the basic kernel is in shape of $H \times W$), and there are $C_o$ augmented convolutional kernels in each layer, leading to the next output tensor with shape of $C_o \times H \times W$. To enable deeper integration of spatial and temporal information, STH divides the input channels into multiple groups and constructs spatio-temporal hybrid kernels by different arrangements of the basic spatial and temporal kernels. Compared to 2D-Conv, STH-Conv can be regarded as replacing a portion of the original spatial convolutional kernels with temporal convolutional kernels in each convolutional layer. A complete network can be built by stacking multiple STH blocks. As such, one resulting advantage of this network is that it maintains an even smaller computational cost and model size than 2D CNNs. Moreover, it enjoys deeper integration of spatial and temporal information. STH block is general and can be applied to existing network architectures such as VGG \cite{vgg}, ResNet \cite{resnet} and MobileNet \cite{mobilenets} to transform them into spatio-temporal networks that can handle spatio-temporal data, e.g., videos.

The main contributions of this work can be summarized as follows:
\begin{itemize}
    \item A novel convolutional block called STH block is proposed to efficiently process spatio-temporal data (e.g., videos). The designed STH block can be applied to any existing architectures such as VGG and ResNet to build complete spatio-temporal networks.
    \item We explore different designs of STH including different integrations of spatial and temporal information, different kernel types, etc. 
    \item We empirically show that the proposed new convolution has competitive spatio-temporal modeling ability while maintaining a small parameter cost.
\end{itemize}

\section{Related Work}
Earlier methods for action recognition (e.g., improved Dense Trajectory (iDT) \cite{dt,idt}) usually use hand-crafted visual features such as Histogram of Gradient (HOG), Histogram of Optical Flow (HOF) \cite{hog}, Motion Boundary Histograms (MBH) \cite{mbh}, SIFT-3D \cite{sift} and Extended SURF \cite{surf} to represent the spatio-temporal information of an action. Yet, hand-crafted representations suffer from the unsatisfactory performance. Recently, deep learning-based methods have been attracting much attention due to their performance superiority over the conventional methods. In the following paragraphs we will briefly review existing deep learning-based methods for action recognition.

\vspace{5pt}
\noindent\textbf{Spatio-Temporal Convolution.} \textit{2D CNNs} \cite{lifeifei,2stream,2streamfusion,tsn,trn} were originally proposed to learn spatial information from static frames. Among these approaches Temporal Segment Network \cite{tsn} could be one of the most widely-used methods for action recognition, in which a sparse sampling strategy is proposed to capture long-term information. However, without sufficient temporal modeling capability, they cannot achieve satisfactory performance on datasets which are sensitive to temporal relationships. Two-stream networks \cite{2stream} alleviate this issue by integrating spatial features captured from video static frames and temporal features captured from extracted optical flow. However, the success of the two-stream methods is heavily relied on the motion features extracted from the offline extracted optical flow, which is expensive to calculate and store.

\vspace{-5pt}
\textit{3D CNNs} \cite{3dcnn} were naturally proposed to jointly encode spatio-temporal information from raw RGB frames. C3D \cite{c3d} network learns spatio-temporal features from a short video sequence (e.g., 16 consecutive frames). I3D \cite{i3d}, one of the representative methods of 3D CNNs, utilizes powerful pre-trained parameters from ImageNet dataset by inflating 2D convolutions to 3D convolutions along the temporal dimension. However, 3D convolution increases computational complexity substantially and greatly affects inference efficiency. Some works \cite{p3d,s3d,r2+1d} factorize the spatial-temporal 3D convolution into one 2D spatial convolution and one 1D temporal convolution, and than ensemble them in a sequential or parallel manner. However, the spatial and temporal information are modeled independently. In addition, memory demand is also increased due to the increase of network layers. LGD \cite{lgd} learns local and global representation by LGD block with diffusion operations. SlowFast Networks \cite{slowfast} capture spatial semantics at low frame rate with a ``slow'' pathway and capture motion at finer temporal resolution with a ``fast'' pathway.

\vspace{5pt}
\noindent\textbf{Efficient Models.} Effective and Efficient models for action recognition have raised much interest among the research community. ECO \cite{eco} adopts 2D convolution at the bottom layers and 3D convolution at the top layers by combining the lower computational cost of 2D convolution with the spatial-temporal modeling ability of 3D convolution. However, it sacrifices the temporal modeling capability at some 2D layers. The latest method TSM (Temporal Shift Module) \cite{tsm} shift part of the channels along the temporal dimension for efficient temporal modeling. STM (Spatio-Temporal and Motion Encoding) \cite{stm} designs two module, specifically, a channel-wise spatio-temporal module to encode spatio-temporal feature and a channel-wise motion module to learn motion feature. CSN (Channel-Separated Convolution Networks) \cite{csn} explores the significance of channel interaction in the network. Our proposed STH-Conv not only has powerful collaborative spatio-temporal modeling capability, but also reduces the computational complexity compared to 2D CNNs, which is more light-weight and efficient.

\section{Method}
In this section we first briefly review the conventional 2D \& 3D convolutions, and then introduce the model details of our proposed Spatio-Temporal Hybrid Convolutions.

\subsection{2D \& 3D Convolutions}
Given an input tensor $\mathbf{I} \in \mathbb{R}^{C \times T \times H \times W}$, where $C$, $T$, $H$, $W$ represent the dimension of channel, time (or \#frame), height and width, respectively. Note that we omit ``batch'' dimension for brevity. The initial input tensor is $\mathbf{I}^{(0)} \in \mathbb{R}^{3 \times T_0 \times H_0 \times W_0}$, as each input frame is usually a three-channel RGB image.

The output tensor of 3D convolution can be formally written as:
\begin{equation}
\vspace{-10pt}
\begin{aligned}
\mathbf{O}_{m,t,h,w} &= \sum_{c,k,i,j}^{C_i,K_T,K_H,K_W} \mathbf{W}_{m,c,k,i,j} \mathbf{I}_{c,t+k,h+i,w+j}\\
&= \sum_{c}^{C_i} \sum_{k,i,j}^{K_T,K_H,K_W} \mathbf{W}_{m,c,k,i,j} \mathbf{I}_{c,t+k,h+i,w+j},
\end{aligned}
\label{eq_3d_conv}
\end{equation}
where m is the index of output channels, and $\mathbf{W} \in \mathbf{R}^{C_o \times C_i \times K_T \times K_H \times K_W}$ is the learnable weight matrix. $K_T \times K_H \times K_W$ is the kernel size of basic convolutions, and $C_o$, $C_i$ are the number of output and input channels. Each 3D convolutional kernel $\mathbf{W}_{m,c}$ exhaustively processes all input channels, as can be evidenced by Eq. \ref{eq_3d_conv}. In one perspective, there are $C_o \times C_i$ \textbf{basic kernels} with shape of $K_T \times K_H \times K_W$. In another perspective, a 3D convolutional layer is composed of $C_o$ \textbf{augmented kernels} with shape of $C_i \times K_T \times K_H \times K_W$. The computational complexity for a 3D convolutional layer is $C_o \times C_i \times T \times H \times W \times K_T \times K_H \times K_W$.

2D convolution can be regarded as a special case of 3D convolution, where $K_T = 1$, meaning that there is no weight connection along the temporal dimension. A 2D convolutional layer can be regarded as $C_o$ augmented kernels with shape of $C_i \times 1 \times K_H \times K_W$. The computational complexity of a conventional 2D convolutional layer is therefore: $C_o \times C_i \times T \times H \times W \times 1 \times K_H \times K_W$.

\begin{figure}[t]
\centering
\includegraphics[width=0.7\textwidth]{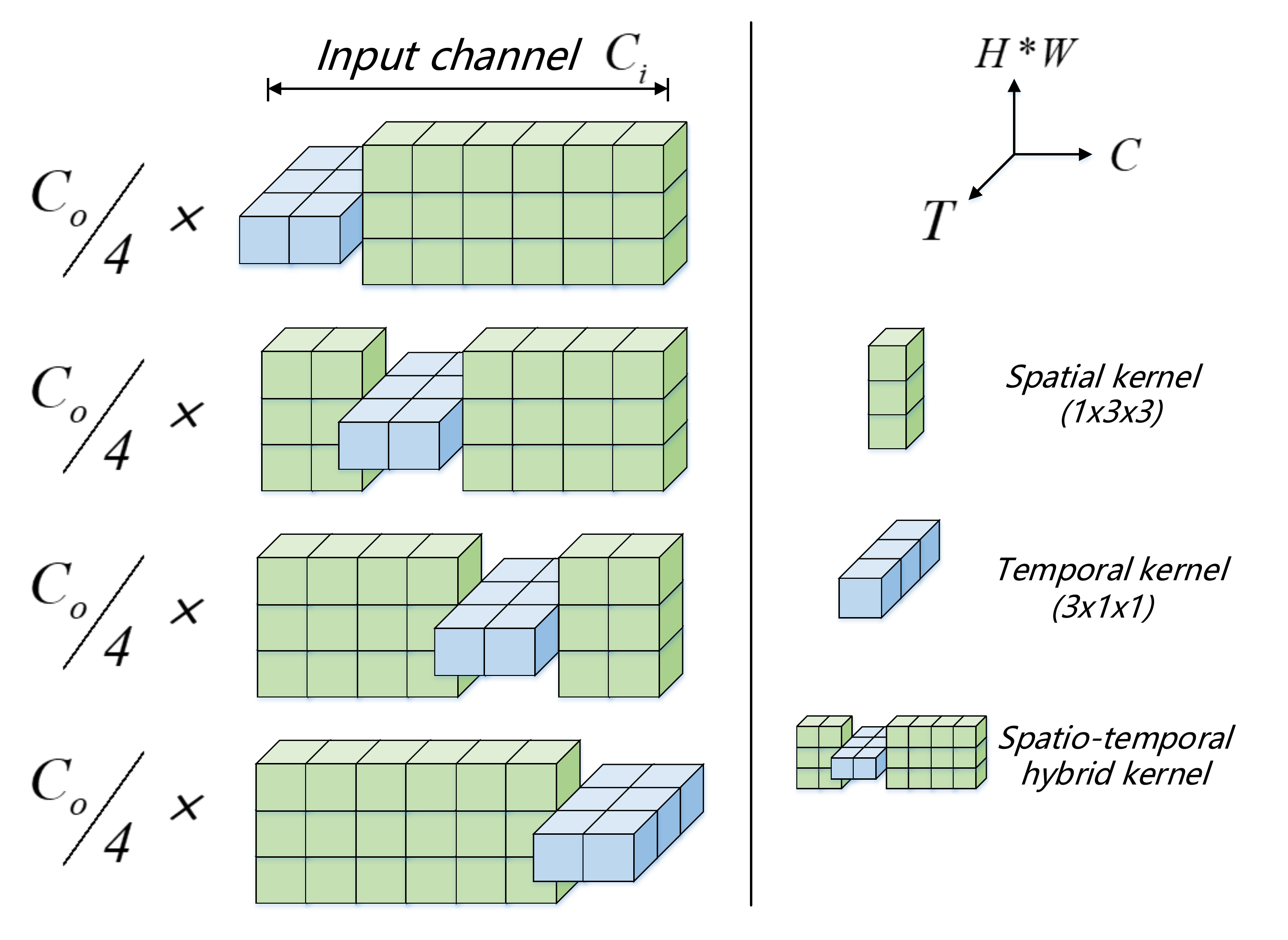}
\caption{Spatio-Temporal Hybrid Convolution (STH-Conv). For visualization, \textbf{the spatial dimensions ($H$ and $W$) are merged as one dimension}. Temporal kernels are shifted along the channel dimension ($C_i = 8$ in this case) to generate multiple combinations of spatio-temporal kernels. Each combination of the spatial and temporal kernels can be treated as one \textbf{hybrid kernel} that spans along the channel dimension. Different hybrid kernels are responsible for extracting spatial and temporal clues corresponding to different input channels. In this figure there are 4 different types of hybrid kernels ($C_o/4$ for each type).}
\vspace{-10pt}
\label{figure_method}
\end{figure}

\subsection{Spatio-Temporal Hybrid Convolution}
Even for the 2D convolution, the $C_o$ augmented convolutional kernels in shape of $C_i \times 1 \times K_H \times K_W$ have been experimentally demonstrated to be redundant \cite{han2015deep}. To reduce the computational cost of 2D convolutions, meanwhile to perform temporal modeling, we construct spatio-temporal hybrid kernels (Figure \ref{figure_method}) by replacing part of spatial kernels with temporal kernels, which endows the convolutions with temporal reasoning capability. We call the proposed method Spatial-Temporal Hybrid Convolution (STH). As illustrated in Figure \ref{figure_method}, along the input channel, a portion (e.g., $1/4$) of the basic spatial convolutional operations ($1 \times K_H \times K_W$) are replaced by temporal convolutional operations ($K_T \times 1 \times 1$). Different replacements result in different spatio-temporal hybrid kernels. Figure \ref{figure_method} shows four different spatio-temporal hybrid kernels that are generated by such replacements. Formally, the output tensor of one STH-Conv can be described by the following equation:
\begin{equation}
\vspace{-5pt}
\begin{aligned}
\mathbf{O}_{m, t, h, w} &=\sum_{c=1}^{p C_{i}} \sum_{k}^{K_T} \mathbf{W}_{m, c, k, 0, 0}^{(T)} \mathbf{I}_{c, t+k, h, w}\\
&+\sum_{c=p C_{i}+1}^{C_{i}} \sum_{i, j}^{K_H,K_W} \mathbf{W}_{m, c, 0, i, j}^{(S)} \mathbf{I}_{c, t, h+i, w+j}\\
&=\mathbf{O}_{m, t, h, w}^{(T)} + \mathbf{O}_{m, t, h, w}^{(S)},
\end{aligned}
\label{eq_sth_conv}
\end{equation}
where $\mathbf{W}_{m,c}^{(T)} \in \mathbb{R}^{K_T \times 1 \times 1}$ is a temporal convolution kernel, $\mathbf{W}_{m,c}^{(S)} \in \mathbb{R}^{1 \times K_H \times K_W}$ is a spatial convolution kernel, $p$ is the proportion of temporal kernels. \textbf{The above equation depicts the first type of hybrid kernel in Figure. \ref{figure_method}}. In order to learn temporal clues more effectively, we diversify $\mathbf{W}_{m}$ ($m=0,1,...,C_o$) by shifting the temporal kernel along input channels, as illustrated in Figure \ref{figure_method}. Each shifting position generates one combination of the spatial and temporal kernels, which can be regarded as one hybrid kernel that spans along the channel dimension. Such a diversity design helps extracting sufficient spatial and temporal clues from multiple channels.

The \#param of an augmented kernel of the 2D-Conv is $C_i \times 1 \times K_H \times K_W$, while the \#param of the corresponding spatio-temporal hybrid kernel of our proposed STH-Conv is $pC_i \times K_T \times 1 \times 1 + (1-p)C_i \times 1 \times K_H \times K_W$. Modern convolutional networks usually adopt the practical suggestion: $K_T = K_H = K_W = 3$. Take $p = 0.5$ as an example, the \#param of a 2D augmented kernel will be $9 \times C_i$, while the \#param of STH-Conv augmented kernel will be $6 \times C_i$. The computational complexity is proportional to the number of convolutional parameters.

\vspace{3pt}
\noindent\textbf{Difference with Existing Modules.} 3D-Conv \cite{3dcnn,c3d,i3d} directly applies 3D convolutional kernels (e.g., $3 \times 3 \times 3$), resulting into a heavy computational cost. (2+1)D-Conv \cite{eco,mixed2d3d,r2+1d,s3d,p3d} decomposes the 3D kernels into 2D spatial kernels and 1D temporal kernels, which are sequentially stacked. Different from the above convolutions which regard all input channels as a whole, STH constructs spatio-temporal hybrid kernels by mixing the basic spatial and temporal kernels along the input channels in one convolutional layer, resulting in deeper integration of spatio-temporal information in one layer. The \#param and computational complexity of STH can be even reduced compared to the plain 2D-Conv. The performance superiority of the proposed STH will be verified in Section \ref{section_experiment}.

\begin{figure}
	\centering
	\subfigure[]{
		\begin{minipage}[t]{1.0\linewidth}
			\centering
			\includegraphics[width=2.2in]{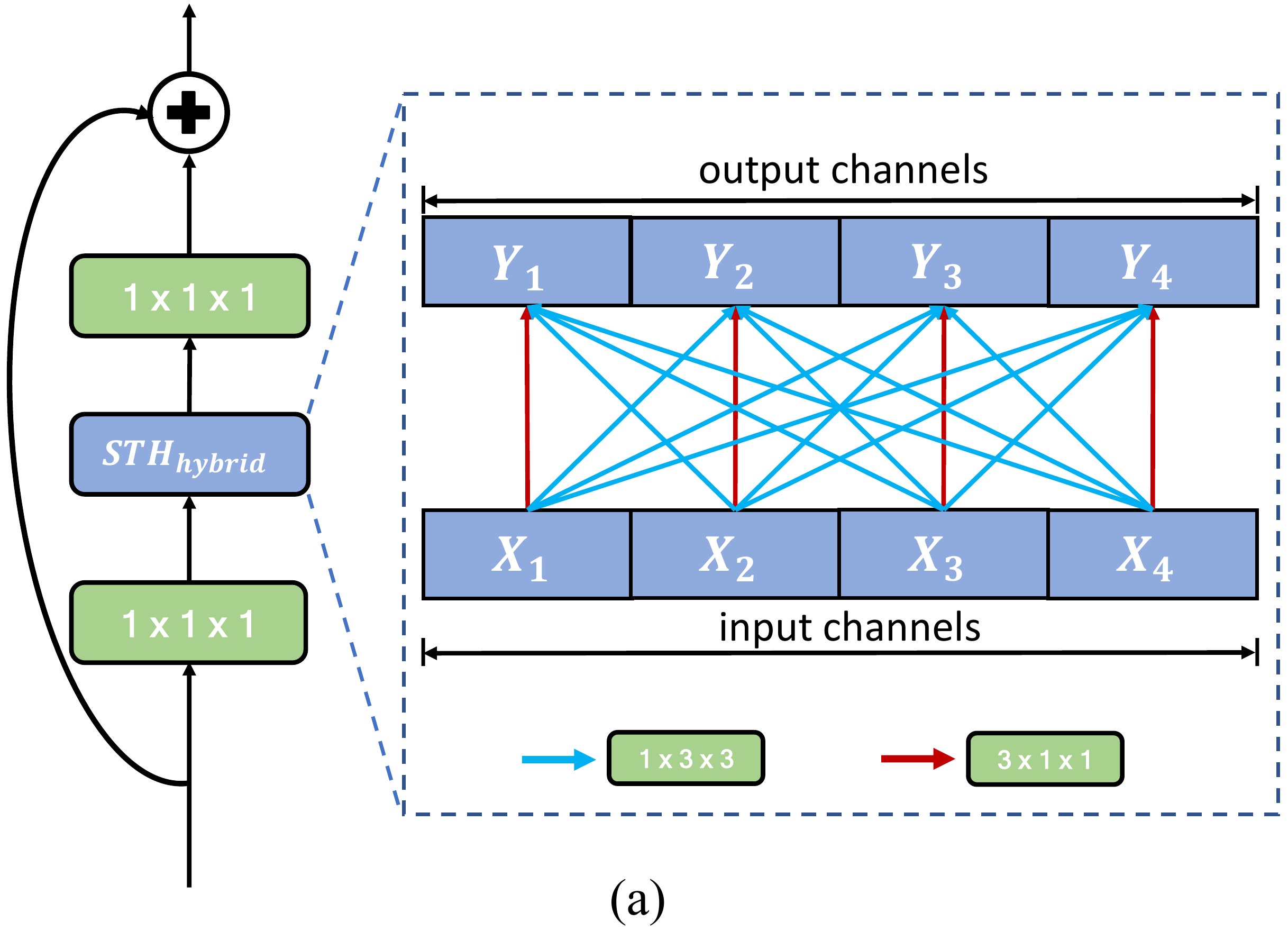}
			\quad\quad
			\includegraphics[width=2.2in]{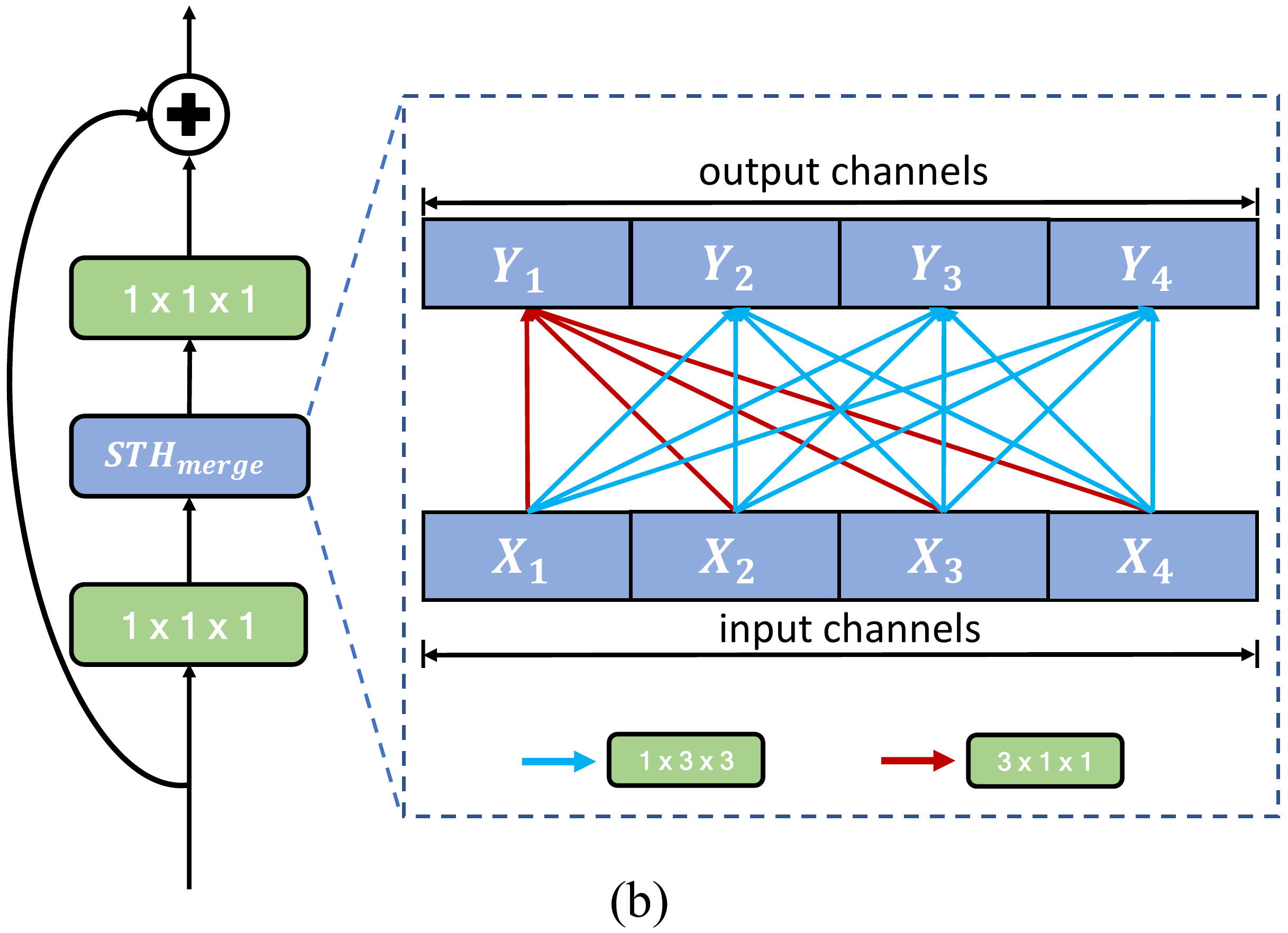}
		\end{minipage}
	}\vskip -15pt
	\centering
	\caption{Overview of two variants of the proposed STH convolution with the proportion of temporal kernels $p = 1/4$. (a) $STH_{hybrid}$. (b) $STH_{merge}$.}
	\vspace{-10pt}
	\label{figure_sthblock}
	\end{figure}
	
\vspace{3pt}
\noindent\textbf{STH Variants.} The STH design in Figure \ref{figure_method} actually corresponds to Figure \ref{figure_sthblock} (a). Figure \ref{figure_sthblock} (b) shows a variant denoted as $STH_{merge}$ without interleaving the spatial and temporal kernels along the input channels.  We will empirically demonstrate in Section \ref{section_ablation} that $STH_{hybrid}$ (a) performs better than $STH_{merge}$ (b) as it could enable deeper integration of the spatial and temporal information from different input channels. Therefore, the following experiments adopt $STH_{hybrid}$ as the default setting.

\subsection{Spatio-Temporal Attentive Integration}
\label{section_attentive}
\begin{figure}
	\centering
	\includegraphics[width=0.9\textwidth]{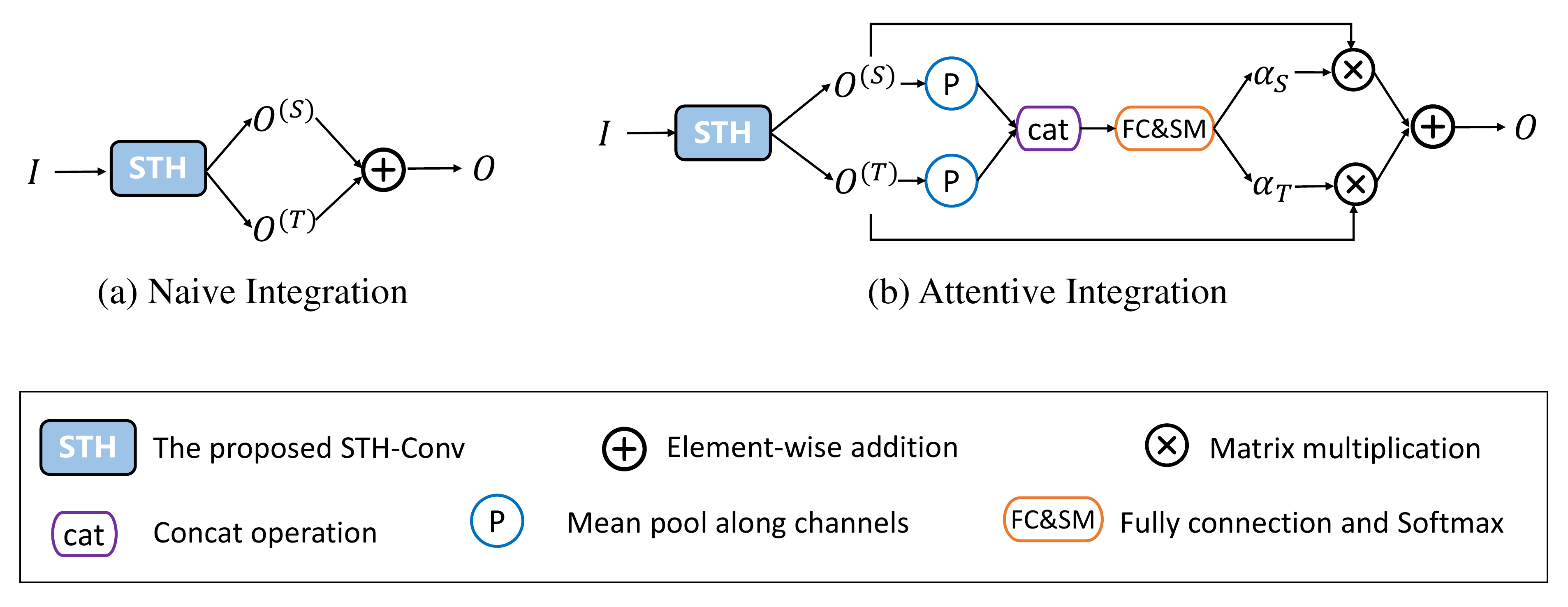}
	\vspace{-5pt}
	\caption{Integration of spatial and temporal features.}
	\vspace{-20pt}
	\label{figure_fusionfig}
	\end{figure}

As shown in Figure \ref{figure_method}, each spatio-temporal hybrid kernel learns to extract useful spatio-temporal clues from multiple input channels. As shown in Eq. \ref{eq_sth_conv}, each spatio-temporal hybrid kernel contains two components, namely, the spatial component $\mathbf{O}_{m, t, h, w}^{(S)}$, and the temporal component $\mathbf{O}_{m, t, h, w}^{(T)}$. One naive method is to directly integrate the two kinds of feature maps with element-wise addition, as in Eq. \ref{eq_sth_conv}. In order to better balance spatial and temporal information, we leverage attention mechanism to adaptively integrate the two kinds of features inside one hybrid spatio-temporal kernel. The attentive integration can be graphically depicted in Figure \ref{figure_fusionfig}. Specifically, the integration can be formally described as:
\begin{equation}
\label{fusion}
\hat{\mathbf{O}}_{m, t, h, w}=\alpha_{T} \mathbf{O}_{m, t, h, w}^{(T)}+\alpha_{S} \mathbf{O}_{m, t, h, w}^{(S)},
\end{equation}
where $\alpha_{T},\alpha_{S} \in \mathbb{R}^{C_o \times 1}$, representing attention coefficients for the temporal and spatial component along the output channels, respectively. $\alpha_{T}$ and $\alpha_{S}$ will be broadcast to other dimensions for $\mathbf{O}^{(T)}$ and $\mathbf{O}^{(S)}$ respectively. The detailed derivation of $\alpha_{T},\alpha_{S}$ is shown in Figure \ref{figure_fusionfig}.

\vspace{-10pt}
	
\subsection{STH Networks}

Similar to existing 2D or 3D CNNs, we additionally stack one $1 \times 1 \times 1$ convolutional block before and after STH-Conv to build a complete STH block. A complete network can be built by stacking multiple STH blocks. The detailed network configuration is shown in Table \ref{table_configure}.

\begin{table}[htb]
\caption{STH Network Architecture.}\smallskip
\vspace{-10pt}
\centering
\resizebox{0.6\columnwidth}{!}{
\smallskip\begin{tabular}[b]{ c | c | c }
\hline
{Layer name} & {Output size ($C$,$T$,$H$,$W$)} & {STH block} \\
\hline
Input & $3\times 8\times 224\times 224$ & -- \\
\hline
Conv1 & $64 \times 8 \times 112 \times 112$ & $1 \times 7 \times 7$, (stride: $2\times 2)$ \\
\hline
Pool1 & $64 \times 8 \times 56 \times 56$ & max, $1\times 3\times 3$, 64 (stride: $2\times 2)$ \\
\hline
$Conv2_x$ & $256\times 8\times 56\times 56$ &
$\left[\begin{array}{c}
1\times 1 \times 1, 256 \\ 
\textit{STH-Conv}, 64 \\ 
1\times 1 \times 1, 256
\end{array}\right] \times 3$ 
\\
\hline
$Conv3_x$ & $512\times 8\times 28\times 28$ & $\left[\begin{array}{c} 1\times 1 \times 1, 512 \\ \textit{STH-Conv}, 128 \\ 1\times 1 \times 1, 512\end{array}\right] \times 4$ \\
\hline
$Conv4_x$ & $1024\times 8\times 14\times 14$ & $\left[\begin{array}{c}1\times 1 \times 1, 1024 \\ \textit{STH-Conv}, 256 \\ 1\times 1 \times 1, 1024\end{array}\right] \times 6$ \\
\hline
$Conv5_x$ & $2048\times 8\times 7\times 7$ & $\left[\begin{array}{c}1\times 1 \times 1, 2048 \\ \textit{STH-Conv}, 512 \\ 1\times 1 \times 1, 2048\end{array}\right] \times 3$ \\
\hline
Pool5 & $2048\times 8\times 1\times 1$ & avg, $1\times 7\times 7$ \\
\hline
FC & $8\times num_{class}$ & $2048 \times num_{class}$ \\
\hline
Consensus & $1\times num_{class}$ & avg\_consensus \\
\hline
\end{tabular}
}
\vspace{-13pt}
\label{table_configure}
\end{table}

\section{Experiments}
\label{section_experiment}
In this section, we first compare STH convolution with different types of convolutions, including 2D-Conv, 3D-Conv, (2+1)D-Conv on Something-Something V1. Afterwards, we compare our method with existing state-of-the-art methods on other datasets. We provide ablation study on the proposed STH convolutional block using Something-Something V1. Finally, we analyze and compare runtime of different approaches. Besides, we provide meticulous discussion of the behaviours of STH networks in our supplementary material.

\subsection{Datasets}
We adopt the following public datasets for evaluation: Something-Something V1 \& V2 \cite{something}, Jester \cite{jester}, Kinetics-400 \cite{i3d}, UCF-101 \cite{ucf101} and HMDB-51 \cite{hmdb51}. We classify these datasets into two categories as follows.


\vspace{3pt}
\noindent\textbf{Temporal-related Datasets.} Something-Something V1 and V2 contain about 108,499 and 220,847 videos of 174 different human-object interaction actions, respectively. Jester contains about 148,092 videos of 27 different human hand gestures, with about 240k videos for training and 20k videos for validation. 

\vspace{3pt}
\noindent\textbf{Appearance-related Datasets.} Kinetics-400 contains about 260k videos with 400 different human action categories, in which 240k videos are for training and 20k for validation. HMDB-51 contains 6,766 videos spanning over 51 categories, while UCF-101 has 13,320 videos spanning over 101 categories. Both datasets consist of short video clips that are temporally trimmed.

Most action categories in the last three datasets can be recognized by only static appearance information. In contrast, the first three datasets are crowd-sourced, in which recognizing the actions from these datasets depends heavily on temporal-related reasoning. Since our approach is proposed for efficient spatio-temporal modeling capability, we pay more attention to the first three temporal-related datasets. Noticeably, we also achieved competitive results on the last three datasets.

\subsection{Implementation Details}
\noindent\textbf{Network Architecture.} STH network is constructed by replacing the $1 \times 3 \times 3$ convolution with the proposed STH convolution. STH is mainly based on ResNet-50 backbone as most of the mentioned models. We directly cite the results from the original paper for compared methods to avoid the possibly tricky reproduction and meanwhile to guarantee the reliability of the comparison.

\vspace{3pt}
\noindent\textbf{Training.} We adopt the same training strategy as TSN \cite{tsn}. We divide an input video into $T$ segments with equal duration and randomly sample one frame from each segment for training. We first fix the size of the input frame as $256 \times 340$, then fixed-size image patches are randomly cropped and horizontally flipped for data augmentation. The cropped regions are finally resized to $224 \times 224$. Therefore, the input volume $\mathbf{I} \in \mathbb{R}^{N \times C \times T \times 224 \times 224}$, where $N$ is batch size, $C$ is the number of input channels, and $T$ is the number of sampled frames per video. In our following experiment, we set $T = 8$ or $16$. Mini-batch SGD with momentum of 0.9 and weight decay of 5e-4 is adopted for optimization. For Something-Something V1 \& V2 and Jester, we train our models for 50 epochs. 
We adopt ImageNet-pretrained weights to initialize the spatial kernels of our STH convolution, while the parameters of the temporal kernels are randomly initialized. Following \cite{tsn}, we use kinetics-400 pre-trained models as initialization when training on UCF-101 and HMDB-51.

\vspace{3pt}
\noindent\textbf{Inference.} For fair comparison, we sample 1 clip per video and use the central $224 \times 224$ crop in Table \ref{table_somethingv1} and Table \ref{table_runtime}. For other tables, we sample 2 clip per video for temporal-related datasets, and 10 clips for appearance-related datasets, following \cite{nonlocal}. 


\begin{table*}[htb]
\caption{Comparison with different convolutions on Something-Something V1 dataset.}\smallskip
\vspace{-15pt}
\centering
\begin{center}
\resizebox{1.0\columnwidth}{!}{
\smallskip\begin{tabular}[b]{c | c | c | c | c | c | c | c | c | c }
\hline
\multicolumn{1}{c|}{Conv-type} &
\multicolumn{1}{c|}{Methods} &
\multicolumn{1}{c|}{Backbone} &
\multicolumn{1}{c|}{Pretrain} &
\multicolumn{1}{c|}{\#Frame} &
\multicolumn{1}{c|}{GFLOPs/V} &
\multicolumn{1}{c|}{\#Param.(M)} &
\multicolumn{1}{c|}{\tabincell{c}{Val\\Top-1}} &
\multicolumn{1}{c|}{\tabincell{c}{Val\\Top-5}} &
\multicolumn{1}{c}{\tabincell{c}{Test\\Top-1}}\\
\hline
\hline
\multirow{8}{*}{2D-Conv} 
& \multirow{2}{*}{TSN \cite{tsn}} & \multirow{2}{*}{ResNet-50} & \multirow{2}{*}{Kinetics}
& 8 & 33 & 24.3 & 19.7 & 46.6 & - \\
&&&& 16 & 65 & 24.3 & 19.9 & 47.3 & - \\
\cline{2-10}
&\multirow{2}{*}{TRN-Mutiscale \cite{trn}} & BNInception & \multirow{2}{*}{ImageNet} & \multirow{2}{*}{8} & 16 & 18.3 & 34.4 & - & 33.6 \\
&&ResNet-50 &&& 33 & 31.8 & 38.9 & 68.1 & - \\
\cline{2-10}
&TSM \cite{tsm} & \multirow{4}{*}{ResNet-50} & \multirow{4}{*}{Kinetics} & 8 & 33 & 24.3 & 45.6 & 74.2 & - \\
&TSM \cite{tsm} &&& 16 & 65 & 24.3 & 47.2 & 77.1 & 46.0 \\
&$TSM_{En}$ \cite{tsm} &&& 24 & 98 & 48.6 & 49.7 & 78.5 & - \\
&$TSM_{RGB+Flow}$ \cite{tsm} &&& 16+16 & & 48.6 & 52.6 & 81.9 & \textbf{50.7} \\
\hline
\multirow{3}{*}{3D-Conv} 
& I3D \cite{nonlocal_gcn} & \multirow{3}{*}{ResNet-50} & \multirow{3}{*}{Kinetics} & \multirow{3}{*}{32$\times$2} & 306 & 28 & 41.6 & 72.2 & - \\
&Non-local I3D \cite{nonlocal_gcn} &&&& 335 & 35.3 & 44.4 & 76 & - \\
&Non-local I3D + GCN \cite{nonlocal_gcn} &&&& 605 & 62.2 & 46.1 & 76.8 & 45 \\
\hline
\multirow{2}{*}{(2+1)D-Conv} 
& S3D-G \cite{s3d} & BNInception & ImageNet & 64 & 142.8 & - & 48.2 & 78.7 & 42.0 \\
& R(2+1)D \cite{r2+1d} & ResNet-34 & Sports1M & 32 & 152 $\times$ 115 & 63.6 & 45.7 & - & - \\
\hline
\multirow{4}{*}{\tabincell{c}{Mixed \\2D\&3D-Conv}}
& \multirow{2}{*}{ECO \cite{eco}} & \multirow{4}{*}{\tabincell{c}{2D BNInception+\\3D ResNet18}} & \multirow{4}{*}{Kinetics} & 8 & 32 & 47.5 & 39.6 & - & - \\
&&&& 16 & 64 & 47.5 & 41.4 & - & - \\
&$ECO_{En}$Lite \cite{eco} &&& 92 & 267 & 150 & 46.4 & - & 42.3 \\
&$ECO_{En}$ $Lite_{RGB+Flow}$ \cite{eco} &&& 92+92 & - & 300 & 49.5 & - & 43.9 \\
\hline
\multirow{2}{*}{\tabincell{c}{Other \\SOTAs}}
& STM \cite{stm} & ResNet-50 & ImageNet & 16$\times$30 & 67$\times$30 & 24.4 & 50.7 & 80.4 & 43.1 \\
& ABM \cite{abm} & ResNet-50 & ImageNet & 16$\times$3 & 106 & - & 46.1 & 74.3 & - \\
\hline
\multirow{4}{*}{\textbf{STH-Conv}} 
& STH & \multirow{3}{*}{ResNet-50} & \multirow{4}{*}{ImageNet} & 8 & 31 & 23.2 & 46.8 & 76.5 & - \\
&STH &&& 16 & 61 & 23.2 & 48.3 & 78.1 & - \\
&$STH_{En}$ &&& 24 & 92 & 46.5 & 50.4 & 80.3 & 45.8 \\
&$STH_{RGB+Flow}$ &&& 16+16 & - & - & \textbf{53.2} & \textbf{82.2} & 48.0 \\
\hline
\end{tabular}
}
\end{center}
\vspace{-30pt}
\label{table_somethingv1}
\end{table*}

\subsection{Comparison with Different Convolutions on Something-Something V1}
The existing methods can be roughly classified into five categories, namely, 2D-Conv, 3D-Conv, (2+1)D-Conv, Mixed 2D\&3D-Conv and other state-of-the-art (SOTA) methods. The performance comparisons of the proposed STH-Conv with other methods is shown in Table \ref{table_somethingv1}.

\vspace{3pt}
\noindent\textbf{2D-Conv based Methods}. TSN \cite{tsn}, the baseline of our method, present poor performance due to the insufficient temporal modeling capability. It can only achieve 19.9\% top-1 accuracy on the temporal-related Something-Something V1 dataset. In comparison, our proposed STH network achieves 27.1\% and 28.4\% absolute top-1 performance improvement for $T$ = 8, 16 (\#frame) respectively. TRN \cite{trn} can learn temporal reasoning relationship from the features of the last layer, but its performance is still inferior to ours. The recently-proposed TSM \cite{tsm} method performs better than other 2D based methods including TRN \cite{trn} and MFNet \cite{mfnet} as it has stronger temporal modeling ability across all levels. Compared to TSM, our proposed STH network achieves new state-of-the-art performance with 46.8\% top-1 accuracy at $T$ = 8 and 48.3\% top-1 accuracy at $T$ = 16, with even lower computational complexity.

\vspace{3pt}
\noindent\textbf{3D-Conv based Methods}. 3D-Conv based methods include I3D \cite{i3d}, Non-local I3D \cite{nonlocal} and the previous state-of-the-art Non-local I3D + GCN \cite{nonlocal_gcn}. Although Non-local I3D + GCN leverages multiple techniques including extra data (MSCOCO), extra spatio-temporal features (I3D), its performance is still inferior to ours. Noticeably, our STH achieves higher accuracy than Non-local I3D + GCN with 8 and 16 frames as input while using 19.5X and 9.3X fewer FLOPs respectively.

\vspace{3pt}
\noindent\textbf{(2+1)D-Conv and Mixed 2D\&3D-Conv}. To reduce the computational complexity of 3D-Conv based approaches, (2+1)D-Conv (P3D \cite{p3d}, S3D \cite{s3d}, and R(2+1)D \cite{r2+1d}) decompose a 3D convolutional operation into one 2D spatial convolution plus one 1D temporal convolution. As shown in Table \ref{table_somethingv1}, STH with 16 frames already outperforms S3D-G with 64 frames, and R(2+1)D with 32 frames. The other is Mixed 2D\&3D-Conv (ECO \cite{eco}), which uses 2D-Conv in early layers and 3D-Conv in deeper layers (bottom-heavy). Compared with ECO ($T$ = 8), our STH ($T$ = 8) achieves 7.2\% higher top-1 accuracy with less FLOPs and only half of the parameters.

\vspace{3pt}
\noindent\textbf{Other SOTA Methods.} We also compare our method with other recently-proposed approaches, including TSM \cite{tsm}, STM \cite{stm}, and ABM \cite{abm}. As Table \ref{table_somethingv1} shows, STH with 8-frame input already outperforms ABM with $(16\times3)$-frame input. Compared with TSM which also focuses on efficient modeling, STH achieves better top-1 validation accuracy with less FLOPs and parameters. To obtain better performance, we follow the good practice of two-stream fusion with 16-frame input. Our model achieves 53.2\% top-1 accuracy and 82.2 top-5 accuracy on the validation set, surpassing other compared methods.

\begin{table*}[htb]
\caption{Comparison with state-of-the-arts on Something-Something V2 and Jester.}\smallskip
\vspace{-15pt}
\centering
\begin{center}
\resizebox{0.9\columnwidth}{!}{
\smallskip\begin{tabular}[b]{ c | c | c | c | c | c | c | c | c }
\hline
\multirow{2}{*}{Methods} &
\multirow{2}{*}{Backbone} &
\multirow{2}{*}{Pretrain} &
\multirow{2}{*}{\#Frame} &
\multirow{2}{*}{FLOPs(G)/Video} &
\multicolumn{2}{c|}{Something v2} &
\multicolumn{2}{c}{Jester} \\
\cline{6-9} 
&&&&& {Val Top-1} & {Val Top-5} & {Val Top-1} & {Val Top-5} \\
\hline
\hline
\multirow{2}{*}{TSN \cite{tsn}} & \multirow{2}{*}{2D ResNet-50} & \multirow{2}{*}{Kinetics} 
& 8$\times$ 30 & 33$\times$ 30 & 30.0 & 60.5 & 83.9 & 99.6 \\
&&& 16$\times$ 30 & 65$\times$ 30 & 30.5 & 61.2 & 84.6 & 99.6 \\
\hline
TRN-Mutiscale \cite{trn} & \multirow{2}{*}{2D BNInception} & \multirow{2}{*}{ImageNet} 
& 25$\times$ 10 & 99$\times$ 10 & 48.8 & 77.6  & 95.3 & - \\
$TRN_{RGB+Flow}$ \cite{trn} &&& - & - & 55.5 & 83.1 & - & - \\
\hline
\multirow{2}{*}{TSM \cite{tsm}} & \multirow{2}{*}{2D ResNet-50} & \multirow{2}{*}{ImageNet} 
& 8 $\times$ 6 & 33$\times$ 6 & 61.2 & 87.1 & \textbf{97.0} & 99.9 \\
&&& 16 $\times$ 6 & 65$\times$ 6 & 63.1 & 88.2 & - & - \\
\hline
STM \cite{stm} & 2D ResNet-50 & ImageNet & 8 $\times$ 30 & 33$\times$ 30 & 62.3 & 88.8 & 96.6 & 99.9 \\ \hline
ABM \cite{abm} & 2D ResNet-34 & ImageNet & 16 $\times$ 3 & 106 & 61.3 & - & - & - \\
\hline
\multirow{2}{*}{\textbf{STH}} & \multirow{2}{*}{2D ResNet-50} & \multirow{2}{*}{ImageNet} 
& 8 $\times$ 6 & 31$\times$ 6 & 61.8 & 88.1 & 96.2 & 99.8 \\
&&& 16$\times$ 6 & 61$\times$ 6 & \textbf{63.6} & \textbf{88.9} & 96.5 & \textbf{99.9} \\
\hline
\end{tabular}
}
\end{center}
\vspace{-30pt}
\label{table_sthv2&jester}
\end{table*}

\subsection{Comparison on Other Datasets}
\vspace{-3pt}
\noindent\textbf{Results on Temporal-related Datasets.} In addition to Something-Something V1 dataset, we also evaluate our STH on Something-Something V2 and Jester dataset. The results compared with other existing state-of-the-art methods are listed in Table \ref{table_sthv2&jester}. Compared to the 2D-Conv models including TSN \cite{tsn} and TRN \cite{trn}, our approach shows clear performance improvement on both datasets. For Something-Something V2 dataset, our model outperforms the current state-of-the-art efficient method TSM \cite{tsm} with fewer FLOPs and parameters, using the same sampling strategy. In addition, our method sampling with only 8 frames performs better than the latest model ABM\cite{abm}. On Jester dataset, our model also achieves competitive performance with TSM \cite{tsm}, ABM \cite{abm}, and STM \cite{stm}.

\vspace{3pt}
\noindent\textbf{Results on Appearance-related Datasets.} Although our model focuses on temporal modeling, we also evaluate it on appearance-related datasets, including Kinetics-400, UCF-101 and HMDB-51. As can be seen from Table \ref{table_kinetics} and Table \ref{table_ucf&hmdb}, the baseline model TSN, although without good temporal modeling ability, can still achieve high performance. It shows that most of the actions in these appearance-related datasets can still be well recognized by using only the appearance information. We use 2D ResNet-50 as our backbone network and only sample 16 RGB frames as input. 
On Kinetics-400 dataset, compared with the state-of-the-art methods in Table \ref{table_kinetics}, our proposed STH achieves a very competitive performance with 91.6\% top-5 accuracy.


We next conduct transfer learning experiments from Kinetics to other datasets with smaller scale, including HMDB-51 and UCF-101, in order to verify the generalization ability of our model. As shown in Table \ref{table_ucf&hmdb}, we achieve near state-of-the-art performance on UCF-101. On HMDB-51, STH with 16-frame input even outperforms I3D and recent methods including TSM\cite{tsm}, ABM \cite{abm} and STM \cite{stm} with lower computational complexity.

\vspace{-15pt}
\begin{table*}[htb]
\caption{Comparison with the state-of-the-arts on Kinetics-400.}\smallskip
\vspace{-20pt}
\centering
\begin{center}
\resizebox{0.7\columnwidth}{!}{
\begin{tabular}{ c | c | c | c }
\hline
{Methods} & {Backbone} & {Val Top-1} & {Val Top-5} \\
\hline
\hline
TSN \cite{tsn} & ResNet-50 & 67.8 & 87.6 \\
\hline
ARTNet \cite{artnet} & ResNet-18  & 69.2 & 88.3 \\
\hline
ECO \cite{eco} & BNInception+ResNet-18  & 70.7 & 89.4 \\
\hline
S3D \cite{s3d} & Inception  & 72.2 & 90.6 \\
\hline
I3D \cite{i3d} & \multirow{2}{*}{3D Inception-v1}  & 71.1 & 89.3 \\
$I3D_{RGB+Flow}$ \cite{i3d} && \textbf{74.2} & 91.3 \\
\hline
StNet \cite{Stnet} & ResNet-101  & 71.4 & - \\
\hline
R(2+1)D \cite{r2+1d} & \multirow{2}{*}{ResNet-34}  & 72.0 & 90.0 \\
$R(2+1)D_{RGB+Flow}$ \cite{r2+1d} && 73.9 & 90.9 \\
\hline
CoST \cite{cost} & ResNet-50  & 74.1 & 91.2 \\
\hline
DynamoNet \cite{dynamonet} & ResNext101  & 66.3 & 86.7 \\
\hline
ip-CSN \cite{csn} & ResNet-50  & 70.8 & - \\
\hline
TSM \cite{tsm} & ResNet-50  & 74.1 & 91.2 \\
\hline
STM \cite{stm} & ResNet-50  & 73.7 & 91.6 \\
\hline
\textbf{STH} & ResNet-50  & 74.1 & \textbf{91.6}\\
\hline
\end{tabular}
}
\end{center}
\vspace{-30pt}
\label{table_kinetics}
\end{table*}

\begin{table}
\caption{Comparison with the state-of-the-arts on UCF-101 and HMDB-51.}
\smallskip
\vspace{-30pt}
\centering
\begin{center}
\resizebox{1.0\columnwidth}{!}{
\smallskip\begin{tabular}[b]{ c | c | c | c | c | c | c }
\hline
\multirow{2}{*}{Methods} &
\multirow{2}{*}{Backbone} &
\multirow{2}{*}{Pretrain} &
\multirow{2}{*}{\#Frame} &
\multirow{2}{*}{FLOPs(G)/Video} &
UCF-101 &
HMDB-51 \\
\cline{6-7} 
&&&&& {Val Top-1} & {Val Top-1} \\
\hline
\hline
C3D \cite{c3d} & 3D VGG-11 & Sports-1M & - & - & 82.3 & 51.6 \\
\hline
STC \cite{STC} & ResNet-101 & Kinetics & 64$\times 3$ & - & 93.7 & 70.5 \\
\hline
TSN \cite{tsn} & 2D ResNet-50 & ImageNet+Kinetics & 16 & 65$\times$30 & 91.7 & 65.1 \\
\hline
ARTNet \cite{artnet} & 3D ResNet-18 & Kinetics & - & - & 94.3 & - \\
\hline
ECO \cite{eco} & 2D BNInception+3D ResNet-18 & Kinetics & 92 & 267  & 94.8 & 72.4 \\
\hline
I3D \cite{i3d} & 3D BNInception & Kinetics & 92 & 267  & 94.8 & 72.4 \\
\hline
R(2+1)D \cite{r2+1d} & ResNet-34 & ImageNet+Kinetics & 32 & 152$\times$115  & \textbf{96.8} & 74.5 \\
\hline
ABM \cite{abm} & 2D ResNet-50 & Kinetics & - & - & 95.1 & 72.7 \\
\hline
TSM \cite{tsm} & 2D ResNet-50 & ImageNet+Kinetics & 16 & 65$\times$30 & 95.9 & 73.5 \\
\hline
STM \cite{stm} & 2D ResNet-50 & ImageNet+Kinetics & 16 & 67$\times$30 & 96.2 & 72.2 \\
\hline
\hline
\textbf{STH} & ResNet-50 & ImageNet+Kinetics & 16 & 61$\times$30 & 96.0 & \textbf{74.8} \\
\hline
\end{tabular}
}
\end{center}
\vspace{-20pt}
\label{table_ucf&hmdb}
\end{table}

\vspace{-10pt}
\subsection{Ablation Study}
\vspace{-5pt}
\label{section_ablation}
All ablation study is conducted on Something-Something V1 with ResNet-50 backbone and 8-frame input. 

\vspace{3pt}
\noindent\textbf{The Proportion of Temporal Convolution $p$.} $p$ is a trade-off hyper-parameter between spatial and temporal modeling capability. As shown in Table \ref{ablation_part}, the computational complexity decreases as $p$ increases, as one basic temporal kernel ($3 \times 1 \times 1$) has fewer parameters than one basic spatial kernel ($1 \times 3 \times 3$). The 2D baseline network ($p = 0$), which encodes no temporal clues, can only produce unpleasant. Setting $p = 1/8$ directly improves the top-1 accuracy from 19.7\% to 46.6\%. From the table we find that $p = 1/4$ is a good trade-off.

\vspace{-10pt}
\begin{table}[htb]
	\caption{Comparison of different coefficient $p$ on Something-Something V1.}\smallskip
	\vspace{-10pt}
	\centering
	\resizebox{0.6\columnwidth}{!}{
	\smallskip\begin{tabular}[b]{ c | c | c | c | c }
		\hline
		{part} & {FLOPS(G)} & {\#Param.(M)} & {Val Top-1} & {Val Top-5} \\
		\hline
		0 & 33.3 & 24.4 & 19.7 & 46.6 \\
		1/8 & 32.2 & 23.1 & 46.6 & 76.8 \\
		1/4 & 30.5 & 22.0 & \textbf{46.8} & \textbf{77.1} \\
		1/2 & 28.5 & 20.3 & 45.1 & 76.0 \\
		\hline
	\end{tabular}
	}
	\label{ablation_part}
	\vspace{-10pt}
\end{table}

\vspace{3pt}
\noindent\textbf{Different Spatio-Temporal Convolutions.} We provide more rigorous comparison by directly replacing our STH block in Table \ref{table_configure} with (2+1)D block. $(2+1)D_{sequential}$ and $(2+1)D_{parallel}$ are constructed by sequentially or parallelly connecting the 2D convolution ($1 \times 3 \times 3$) with the 1D convolution ($3 \times 1 \times 1$). As in Table \ref{table_spatiotemporal}, STH outperforms both $(2+1)D_{sequential}$ and $(2+1)D_{parallel}$ with a clear margin while maintaining smaller parameter and computational cost. We also compare the different variants of STH (Figure \ref{figure_sthblock}). $STH_{merge}$ performs inferiorly to the hybrid spatio-temporal kernels ($STH_{hybrid}$), which indicates that the hybrid design deeply integrates spatial and temporal clues for better spatio-temporal modeling. Further, we also compare STH convolution ($STH_{hybrid}$) with its attentive integration version (named as $STH_{hybrid+attention}$, illustrated in Section \ref{section_attentive}). With small parameter overhead, $STH_{hybrid+attention}$ outperforms the conventional element-wise addition.

\begin{table}
	\caption{Comparison of different spatio-temporal convolutions on Something-Something V1.}\smallskip
	\vspace{-10pt}
	\centering
	\resizebox{0.7\columnwidth}{!}{
	\smallskip\begin{tabular}[b]{ c | c | c | c | c }
		\hline
		{model} & {FLOPS(G)} & {\#Param.(M)} & {Val Top-1} & {Val Top-5} \\
		\hline
		$(2+1)D_{sequential}$ & 37.9 & 27.6 & 42.7 & 72.5 \\
		$(2+1)D_{parallel}$ & 37.9 & 27.6 & 45.4 & 76.3 \\
		\hline
		$STH_{merge}$ & 30.5 & 22.0 & 45.9 & 76.5 \\
		$STH_{hybrid}$ & 30.5 & 22.0 & 46.8 & 77.1 \\
		$STH_{hybrid+attention}$ & 30.5 & 23.2 & \textbf{47.8} & \textbf{77.9} \\
		\hline
	\end{tabular}
	}
	\label{table_spatiotemporal}
	
\end{table}



\vspace{3pt}
\noindent\textbf{Fixed Kernel Type \textit{vs.} Dilated Kernel Type.} In order to learn multi-scale spatial-temporal information, we also perform ablation study on different convolution kernels. \emph{Fixed kernel type} denotes convolution kernels with dilation rate 1. \emph{Dilated kernel type} denotes convolution kernels with dilation rate $d \in{[1,2,3]}$. The different dilation rates are applied in different layers in an alternating fashion. As shown in Table \ref{ablation_kernel}, dilated kernel type achieves slightly better performance.

\begin{table}[t]
\caption{Ablation study of kernel types on Something-Something V1.}\smallskip
\vspace{-10pt}
\centering
\resizebox{0.6\columnwidth}{!}{
\smallskip\begin{tabular}[b]{ c | c | c | c | c }
\hline
{kernel\_type} & {FLOPS(G)} & {\# Param.(M)} & {Val Top-1} & {Val Top-5} \\
\hline
Fixed & 30.5 & 23.2 & 47.8 & 77.9 \\
Dilated & 30.5 & 23.2 & \textbf{48.4} & \textbf{78.2} \\
\hline
\end{tabular}
}
\label{ablation_kernel}
\vspace{-10pt}
\end{table}

\subsection{Runtime Analysis}
We evaluate the inference runtime of the proposed STH on a single GTX 1080TI GPU. For fair comparison, we use a batch size of 16 and sample 8-frame or 16-frame as input per video. Compared with I3D, our STH achieves 10.7x higher speed (68.8 videos per second \emph{vs.} 6.4 videos per second). Compared to ECO, our model achieves 1.5x higher speed and 5.4\% better accuracy using only half parameters. Compared to TSM and ABM, STH is more favorable regarding both performance and model size. Regarding the inference speed, since the proposed STH-Conv uses hybrid kernels (heterogeneous kernels), it currently obtains limited CUDA support. Further acceleration can be expected in future. 

\vspace{-10pt}
\begin{table}[htb]
	\caption{Runtime and accuracy on Something-Something V1.}\smallskip
	\vspace{-10pt}
	\centering
	\resizebox{0.7\columnwidth}{!}{
		\smallskip\begin{tabular}[b]{ c | c | c | c | c | c}
			\hline
			{Model} & {\#Frame} & {FLOPs(G)} & {\#Param.(M)} & {Speed(VPS)} & {Val Top-1} \\
			\hline
			\hline
			I3D \cite{i3d} & 64 & 306 &  35.3 & 6.4 & 41.6 \\
			\hline
			ECO \cite{eco} & 16 & 64 & 47.5 & 46.3 & 41.4 \\
			\hline
			\multirow{2}{*}{TSM \cite{tsm}}
			& 8 & 33 & 24.3 & 80.4 & 45.6 \\
			& 16 & 65 & 24.3 & 39.5 & 47.2 \\
			\hline
			{ABM-C-$in$ \cite{abm}}
			& $8 \times 3$ & 53.0 & - & 28.0 & 44.1 \\
			{ABM-A-$in$ \cite{abm}}
			& $16 \times 3$ & 106.0 & - & 20.1 & 46.1 \\
			\hline
			\hline
			\multirow{2}{*}{STH}
			& 8 & 30.6 & 23.2 & 68.8 & 46.8 \\
			& 16 & 61.3 & 23.2 & 35.2 & 48.3 \\
			\hline
		\end{tabular}
	}
	\vspace{-25pt}
	\label{table_runtime}
\end{table}

\section{Conclusion}

In this paper, we proposed a novel spatio-temporal convolution, dubbed as STH-Conv, that contains a series of spatio-temporal hybrid kernels to encode spatio-temporal features collaboratively. The proposed method achieves promising performances on benchmark datasets, with even lower computational complexity than the conventional 2D convolution. Specifically the proposed STH-Conv enjoys competitive performance and a favourable parameter cost compared to 2D, 3D, and (2+1)D convolutions.

\vspace{10pt}
\noindent\textbf{Acknowledgement} This work is supported by the National Key Research and Development Program of China under Grant No. 2017YFA0700800.

\vspace{-3pt}
\section{Supplementary Material}
\vspace{-10pt}
\subsection{Analysis of Learnt Spatial and Temporal Attention}
\begin{figure*}[htbp!]
\centering
\vspace{5pt}
\includegraphics[width=0.85\textwidth]{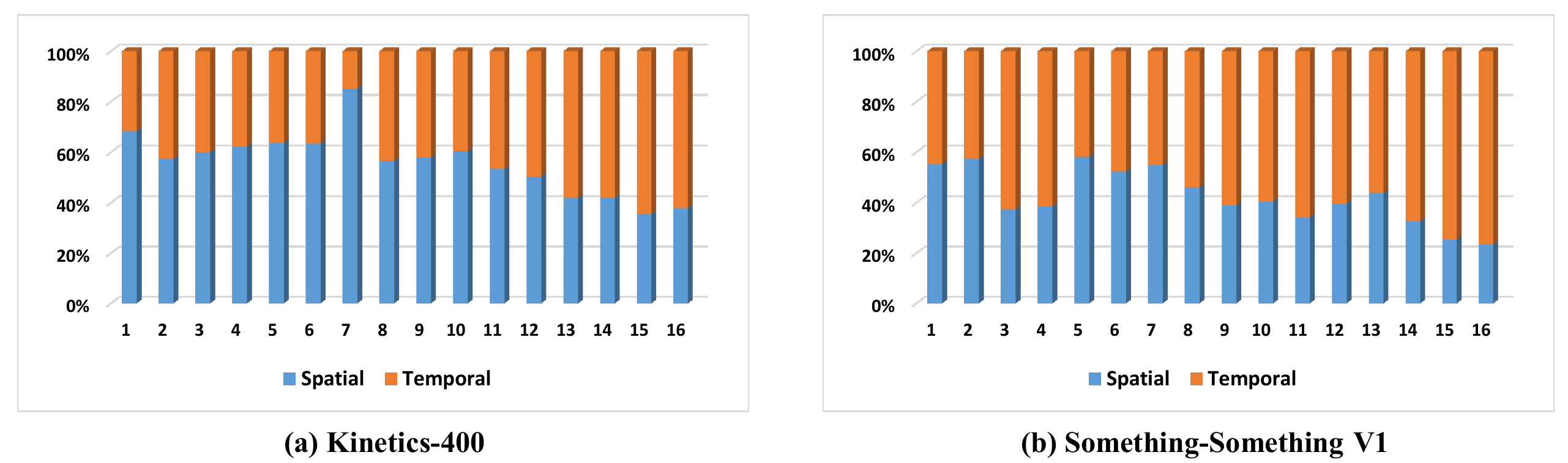}
\caption{The attention coefficients for different convolutional layers.}
\label{figure_layer}
\vspace{-10pt}
\end{figure*}

\begin{figure*}[htbp!]
\centering
\vspace{-5pt}
\includegraphics[width=0.85\textwidth]{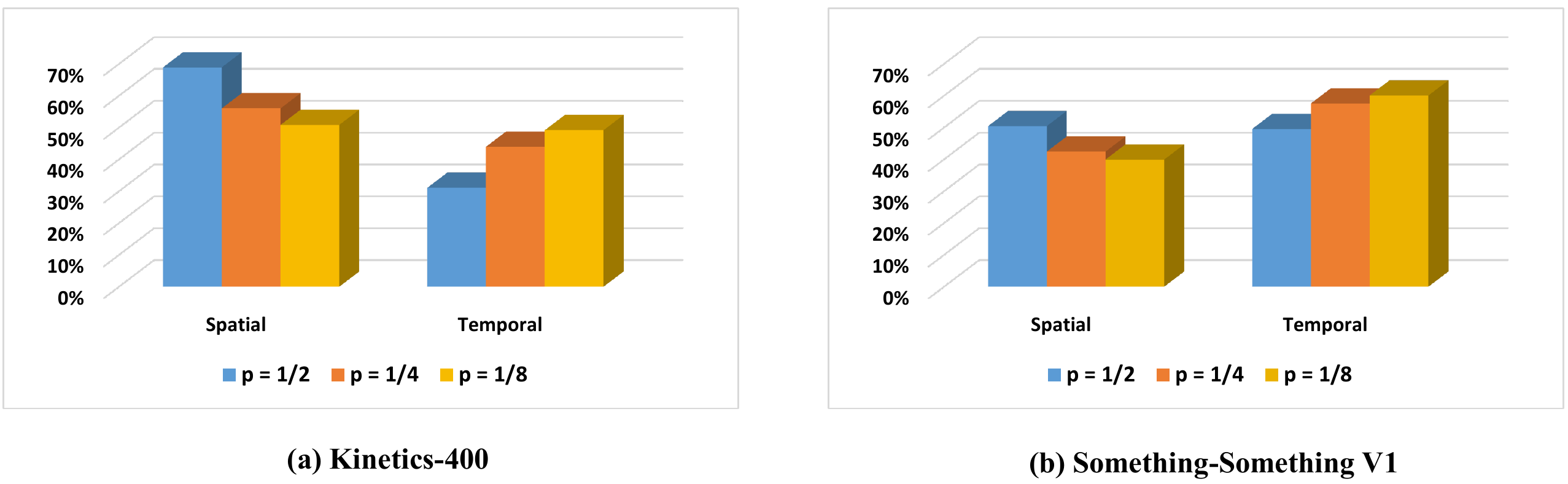}
\caption{The attention coefficients for different temporal proportions $p$.}
\label{figure_proportion}
\vspace{-10pt}
\end{figure*}

\begin{figure*}[htbp!]
\centering
\vspace{-5pt}
\includegraphics[width=0.5\textwidth]{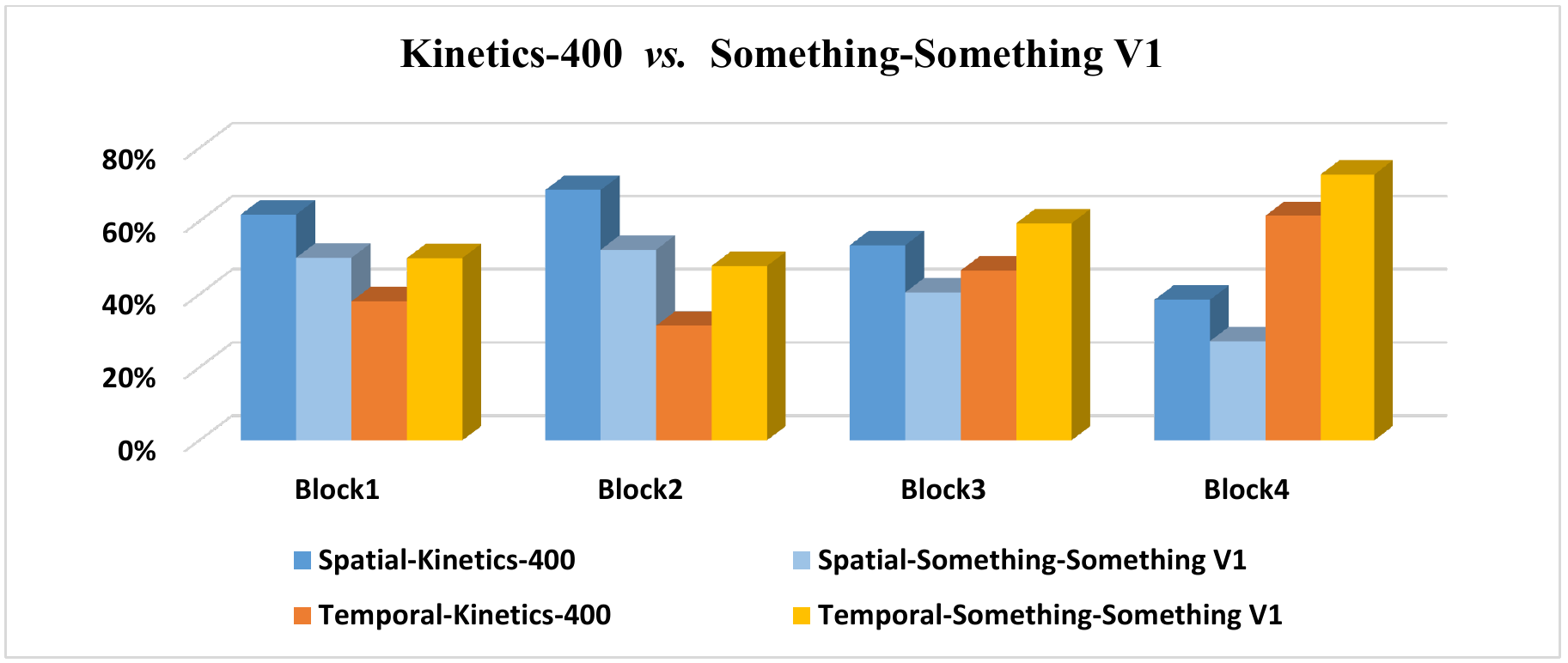}
\caption{The attention coefficients for different datasets.}
\label{figure_dataset}
\vspace{-10pt}
\end{figure*}

\begin{figure*}[htbp!]
\centering
\vspace{-5pt}
\includegraphics[width=0.5\textwidth]{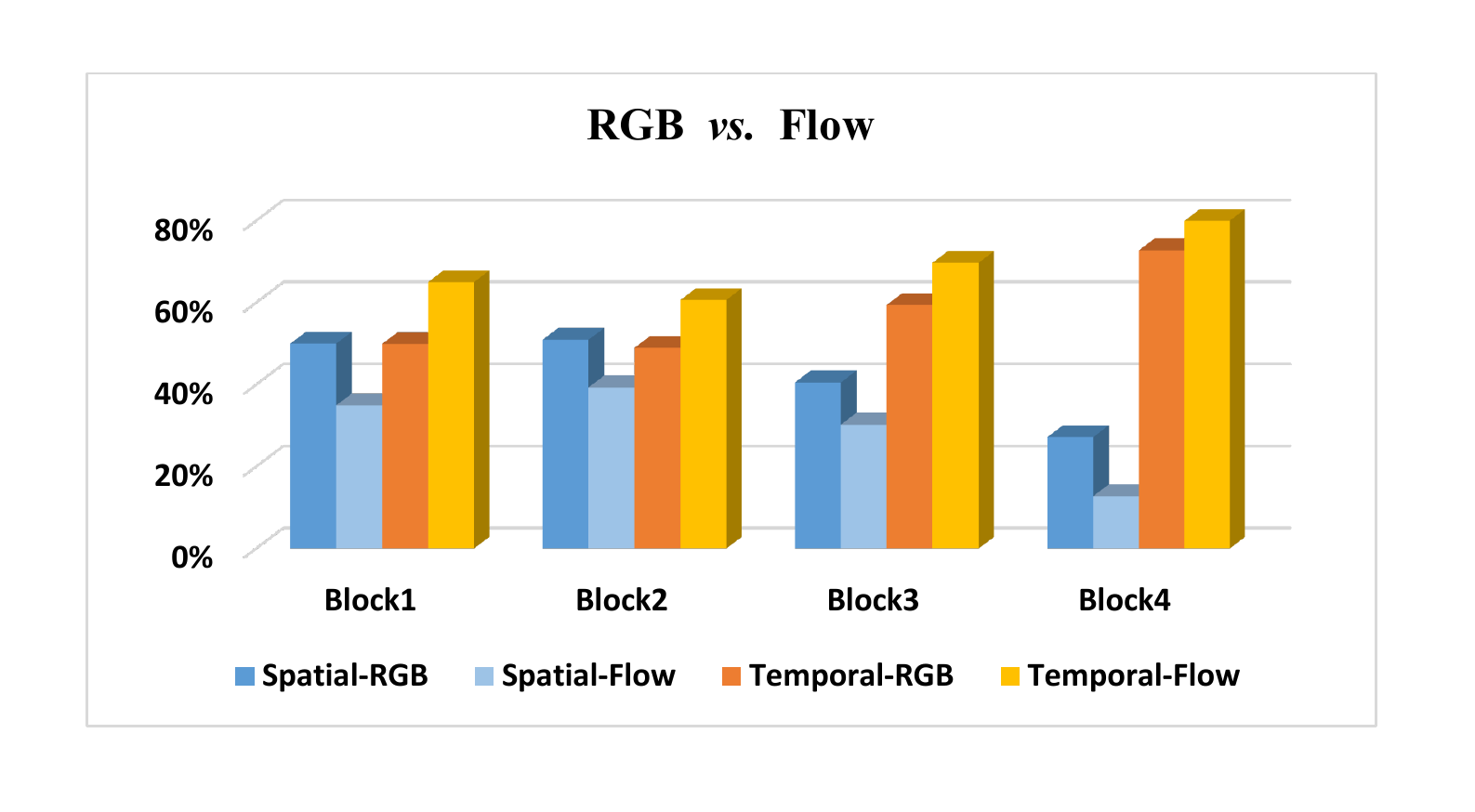}
\caption{The attention coefficients for different modalities (RGB and Flow).}
\label{figure_modality}
\vspace{-10pt}
\end{figure*}

\vspace{-5pt}
As introduced in our main paper, our proposed STH network adaptively integrates the spatial and temporal components with attention mechanism. To uncover the learnt knowledge inside the ``black box'' and better understand the behaviour of the network, we calculate the spatial and temporal attention coefficients for different network layers (or blocks), different portions of temporal components (kernels), different datasets, and different modalities (RGB or Flow). For each image, we calculate the learnt attention coefficient value by averaging the coefficient vectors ($\alpha_{S}$ and $\alpha_{T}$) of different layers for both spatial and temporal components. We average the attention for all images in a dataset. The corresponding results are shown in Figure \ref{figure_layer}, \ref{figure_proportion}, \ref{figure_dataset} and \ref{figure_modality}, respectively.

\vspace{-2.5pt}
\noindent\textbf{Attention for Different Layers.} As can be seen from Figure \ref{figure_layer}, temporal information generally becomes more important for deeper layers. This observation is confirmed across different datasets, including appearance-related datasets (e.g., Kinetics-400) and temporal-related datasets (e.g., Something-Something V1). The possible reasons could be as follows. 1) Most of the useful temporal information (for action classification) is obtained from large spatial receptive field, which relies on deeper layers. 2) The features learnt from shallow layers could be helpful for spatial modeling, but may not be robust to video jitter caused by camera shaking, which could pose negative effect on learning useful temporal clues.

\vspace{-3pt}
\noindent\textbf{Attention for Different Temporal Proportions.} We calculate the attention coefficients for different temporal proportions ($p = 1/2, 1/4, 1/8$) as shown in Figure \ref{figure_proportion}. Although $p$ is a predefined hyper-parameter, our STH network, however, adaptively adjusts attention coefficients to balance the spatial and temporal information. For example, although there is fewer temporal kernels when $p = 1/4$ compared to $p = 1/2$, STH network learns to assign higher attention weights for the temporal information. This observation holds for both Kinetics-400 and Something-Something V1 datasets.

\begin{figure*}[htbp!]
\centering
\vspace{5pt}
\subfigcapskip -2pt
\subfigure[brushing teeth]{
    \begin{minipage}[t]{1.0\linewidth}
        \centering
        \includegraphics[width=0.5in]{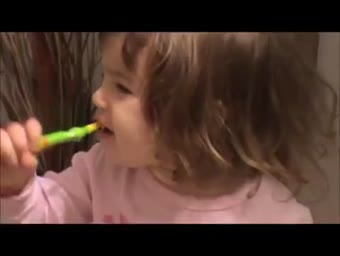}
        \includegraphics[width=0.5in]{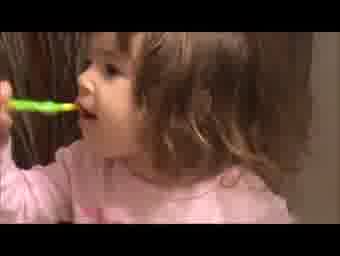}
        \includegraphics[width=0.5in]{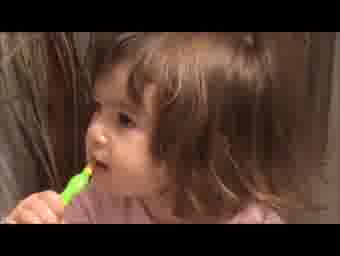}
        \includegraphics[width=0.5in]{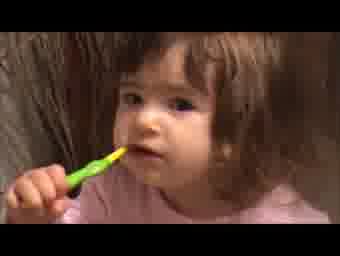}
        \includegraphics[width=0.5in]{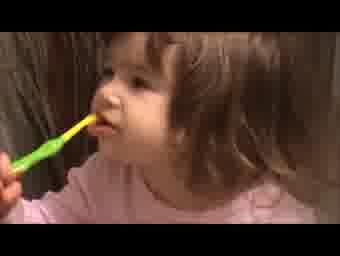}
        \includegraphics[width=0.5in]{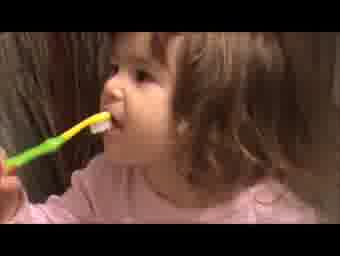}
        \includegraphics[width=0.5in]{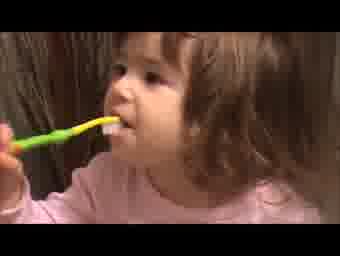}
        \includegraphics[width=0.5in]{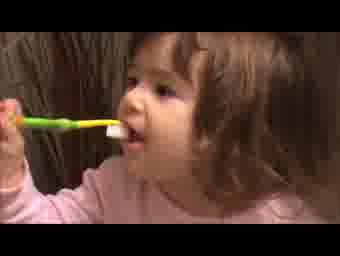}
    \end{minipage}
}\vskip -5pt
\subfigure[carrying baby]{
    \begin{minipage}[t]{1.0\linewidth}
        \centering
        \includegraphics[width=0.5in]{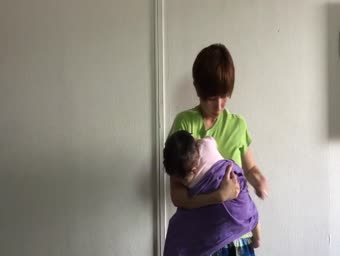}
        \includegraphics[width=0.5in]{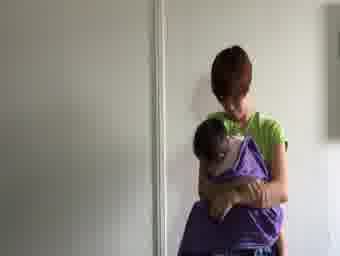}
        \includegraphics[width=0.5in]{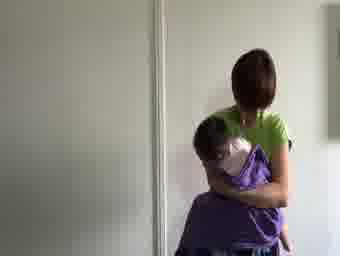}
        \includegraphics[width=0.5in]{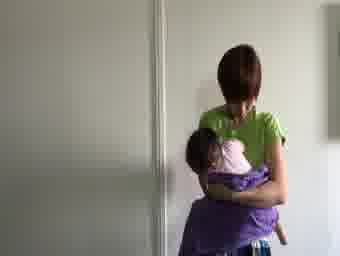}
        \includegraphics[width=0.5in]{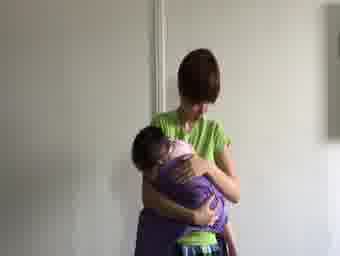}
        \includegraphics[width=0.5in]{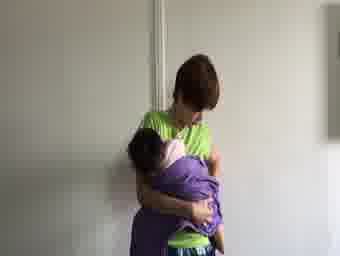}
        \includegraphics[width=0.5in]{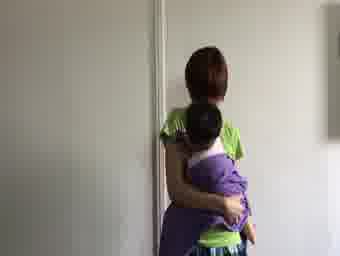}
        \includegraphics[width=0.5in]{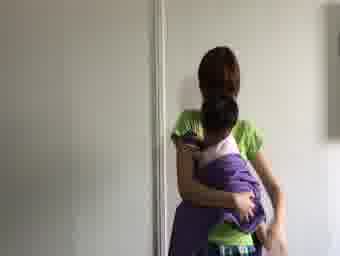}
    \end{minipage}
}\vskip -5pt
\subfigure[playing accordion]{
    \begin{minipage}[t]{1.0\linewidth}
        \centering
        \includegraphics[width=0.5in]{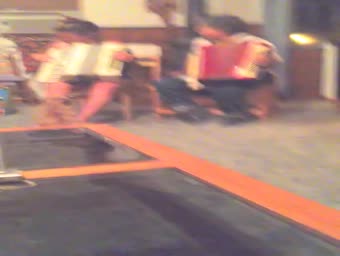}
        \includegraphics[width=0.5in]{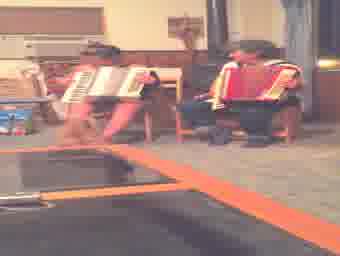}
        \includegraphics[width=0.5in]{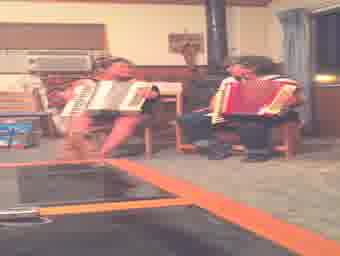}
        \includegraphics[width=0.5in]{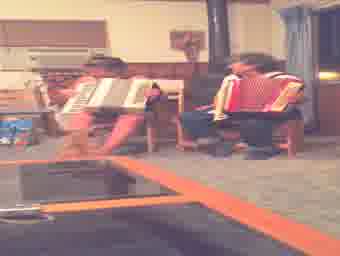}
        \includegraphics[width=0.5in]{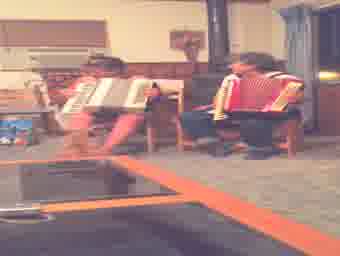}
        \includegraphics[width=0.5in]{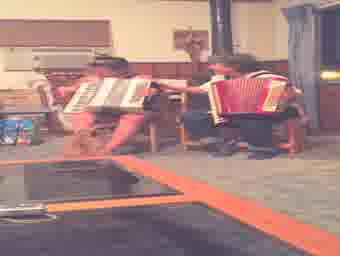}
        \includegraphics[width=0.5in]{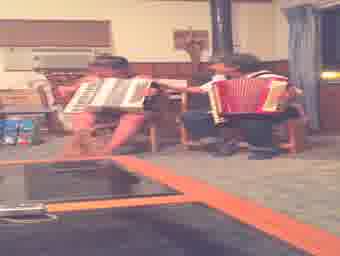}
        \includegraphics[width=0.5in]{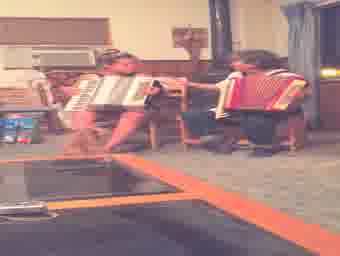}
    \end{minipage}
}\vskip -5pt
\subfigure[eating watermelon]{
    \begin{minipage}[t]{1.0\linewidth}
        \centering
        \includegraphics[width=0.5in]{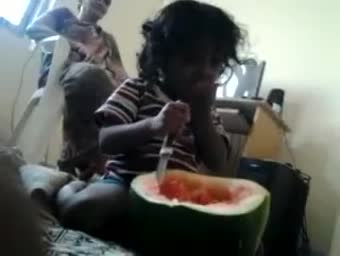}
        \includegraphics[width=0.5in]{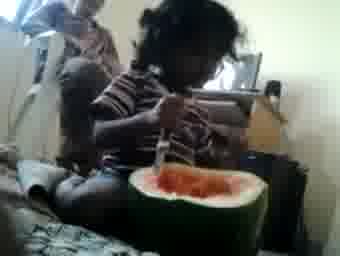}
        \includegraphics[width=0.5in]{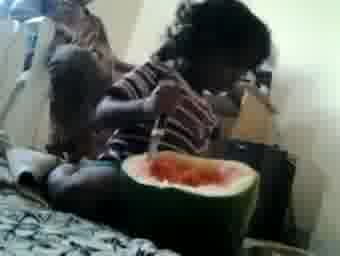}
        \includegraphics[width=0.5in]{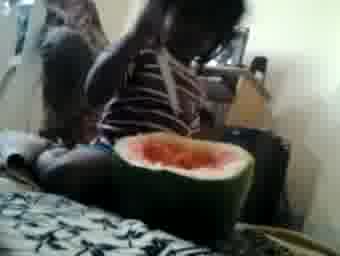}
        \includegraphics[width=0.5in]{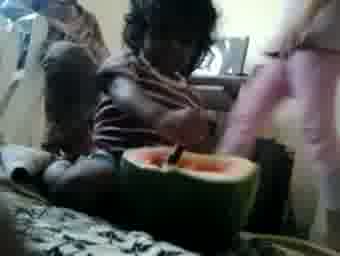}
        \includegraphics[width=0.5in]{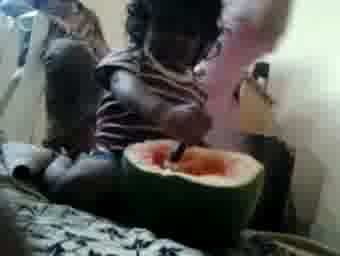}
        \includegraphics[width=0.5in]{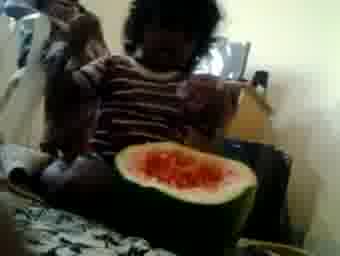}
        \includegraphics[width=0.5in]{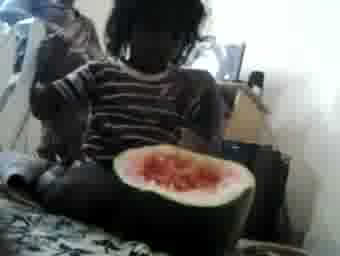}
    \end{minipage}
}\vskip -5pt
\subfigure[getting a tattoo]{
    \begin{minipage}[t]{1.0\linewidth}
        \centering
        \includegraphics[width=0.5in]{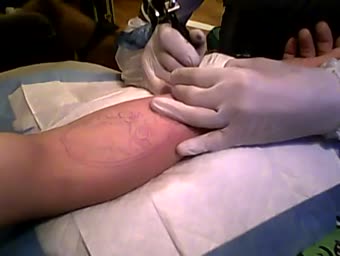}
        \includegraphics[width=0.5in]{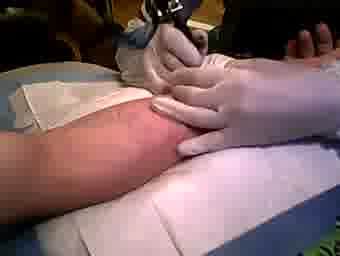}
        \includegraphics[width=0.5in]{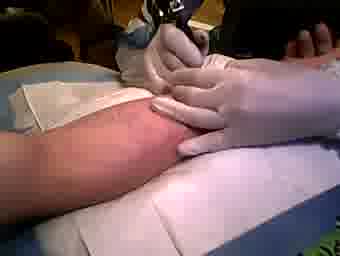}
        \includegraphics[width=0.5in]{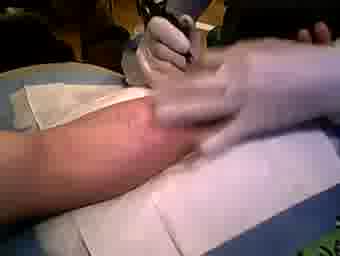}
        \includegraphics[width=0.5in]{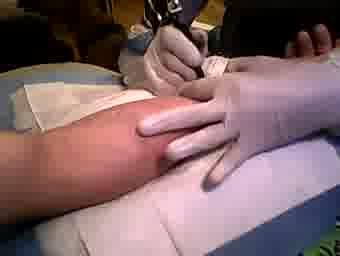}
        \includegraphics[width=0.5in]{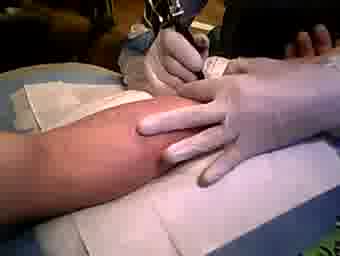}
        \includegraphics[width=0.5in]{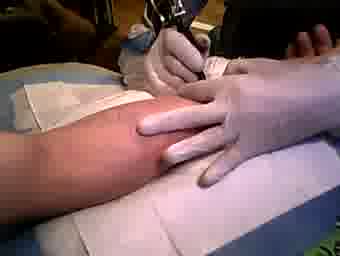}
        \includegraphics[width=0.5in]{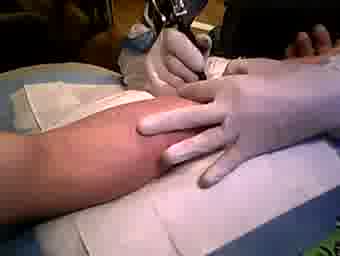}
    \end{minipage}
}\vskip -5pt
\subfigure[running on treadmill]{
    \begin{minipage}[t]{1.0\linewidth}
        \centering
        \includegraphics[width=0.5in]{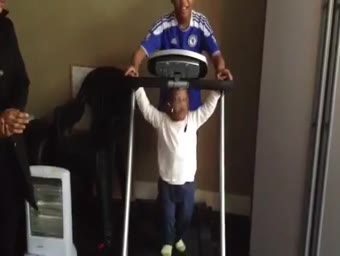}
        \includegraphics[width=0.5in]{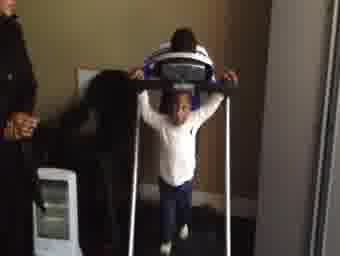}
        \includegraphics[width=0.5in]{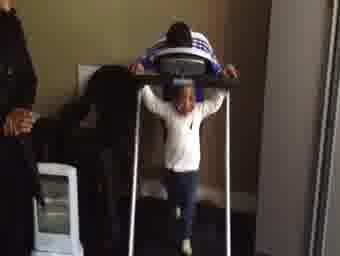}
        \includegraphics[width=0.5in]{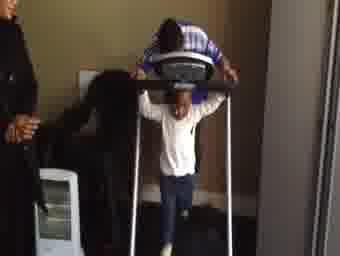}
        \includegraphics[width=0.5in]{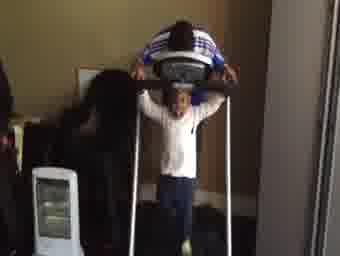}
        \includegraphics[width=0.5in]{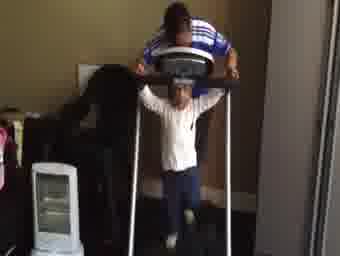}
        \includegraphics[width=0.5in]{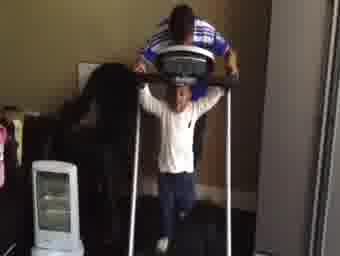}
        \includegraphics[width=0.5in]{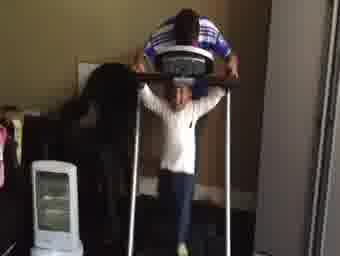}
    \end{minipage}
}
\centering
\caption{Appearance-related categories on Kinetics-400.}
\label{figure_appearance_related}
\vspace{-5pt}
\end{figure*}

\begin{figure*}[htbp!]
\centering
\subfigcapskip -2pt
\subfigure[letting something roll down a slanted surface]{
    \begin{minipage}[t]{1.0\linewidth}
        \centering
        \includegraphics[width=0.5in]{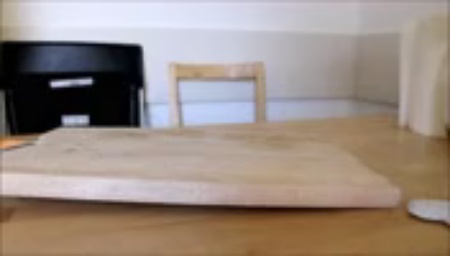}
        \includegraphics[width=0.5in]{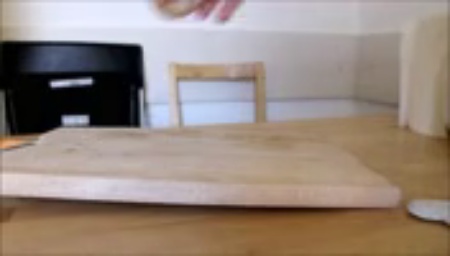}
        \includegraphics[width=0.5in]{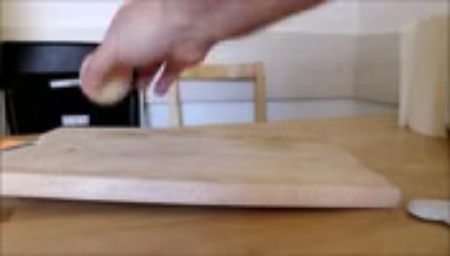}
        \includegraphics[width=0.5in]{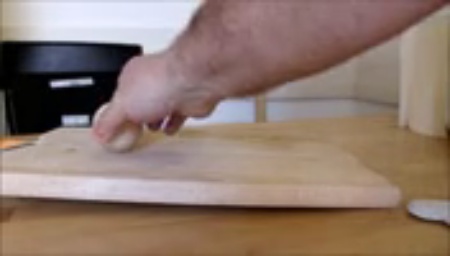}
        \includegraphics[width=0.5in]{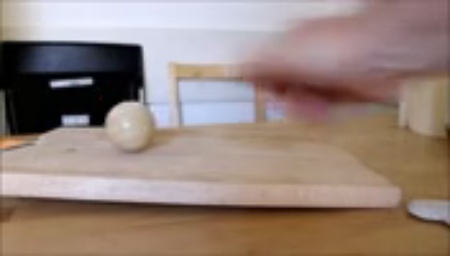}
        \includegraphics[width=0.5in]{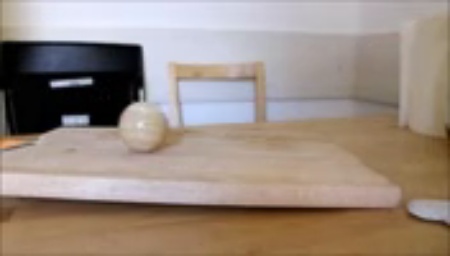}
        \includegraphics[width=0.5in]{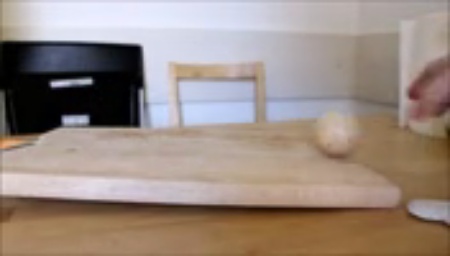}
        \includegraphics[width=0.5in]{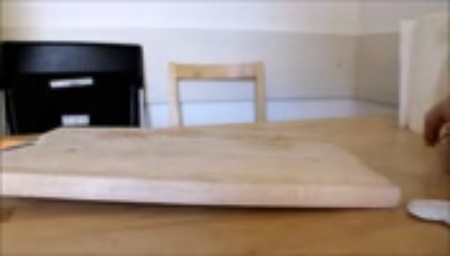}
    \end{minipage}
}\vskip -5pt
\subfigure[moving something and something so they collide with each-other]{
    \begin{minipage}[t]{1.0\linewidth}
        \centering
        \includegraphics[width=0.5in]{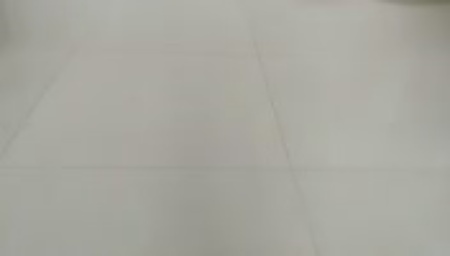}
        \includegraphics[width=0.5in]{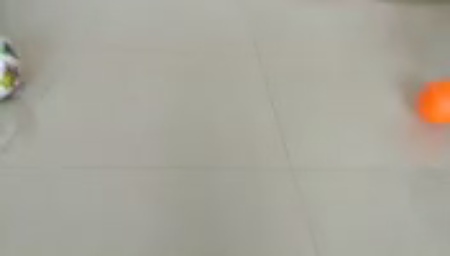}
        \includegraphics[width=0.5in]{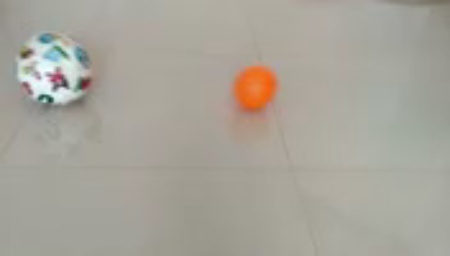}
        \includegraphics[width=0.5in]{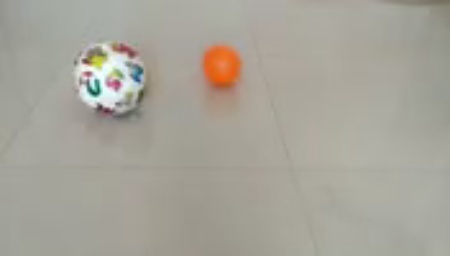}
        \includegraphics[width=0.5in]{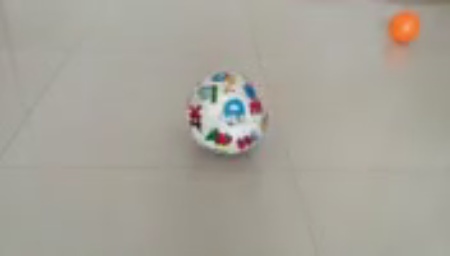}
        \includegraphics[width=0.5in]{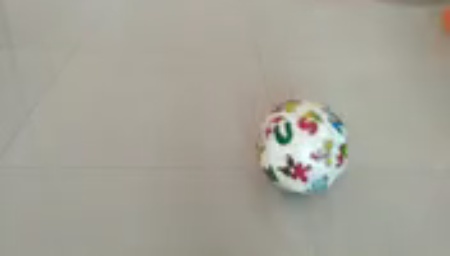}
        \includegraphics[width=0.5in]{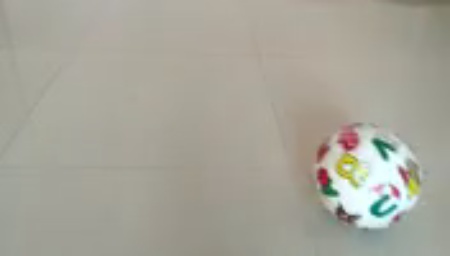}
        \includegraphics[width=0.5in]{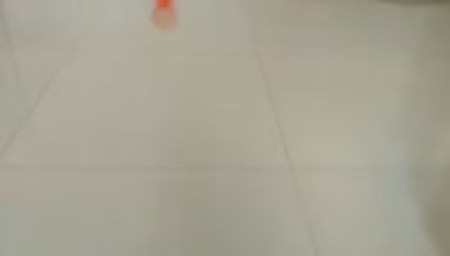}
    \end{minipage}
}\vskip -5pt
\subfigure[moving something and something so they collide with each-other]{
    \begin{minipage}[t]{1.0\linewidth}
        \centering
        \includegraphics[width=0.5in]{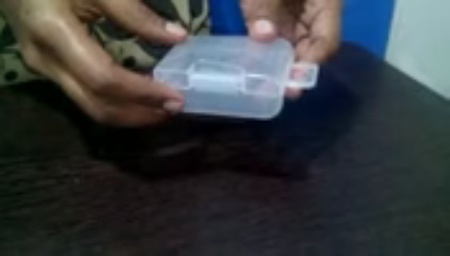}
        \includegraphics[width=0.5in]{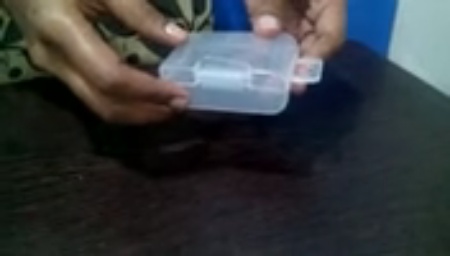}
        \includegraphics[width=0.5in]{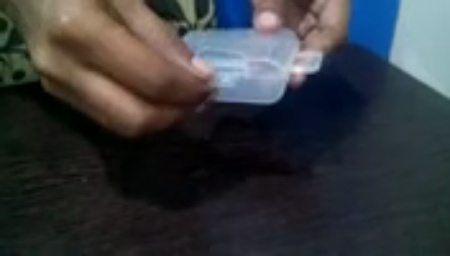}
        \includegraphics[width=0.5in]{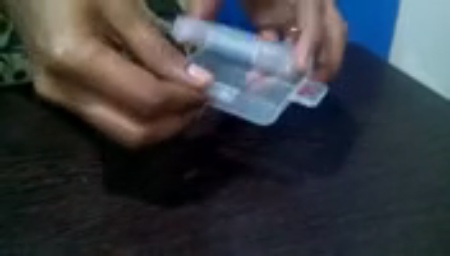}
        \includegraphics[width=0.5in]{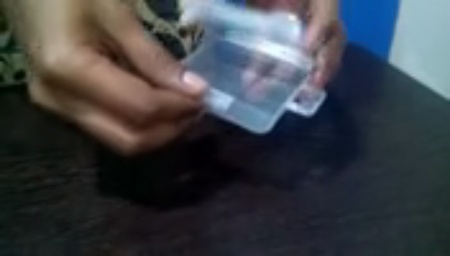}
        \includegraphics[width=0.5in]{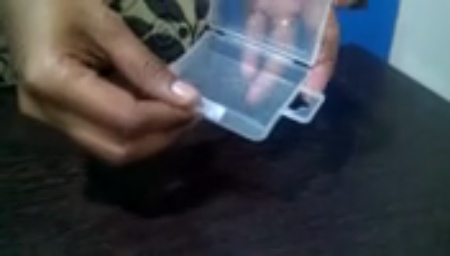}
        \includegraphics[width=0.5in]{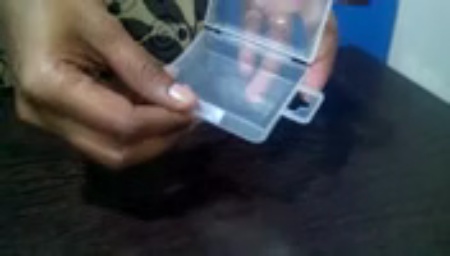}
        \includegraphics[width=0.5in]{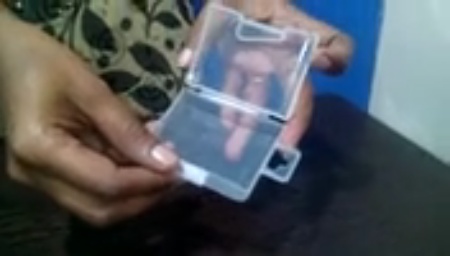}
    \end{minipage}
}\vskip -5pt
\subfigure[poking a stack of something so the stack collapses]{
    \begin{minipage}[t]{1.0\linewidth}
        \centering
        \includegraphics[width=0.5in]{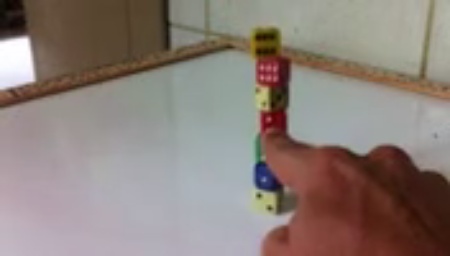}
        \includegraphics[width=0.5in]{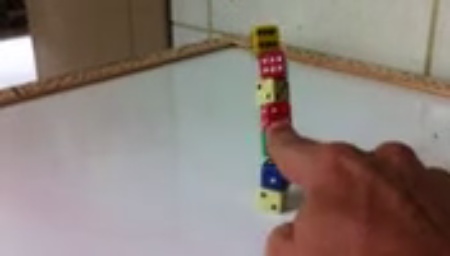}
        \includegraphics[width=0.5in]{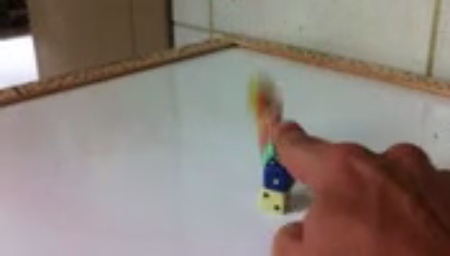}
        \includegraphics[width=0.5in]{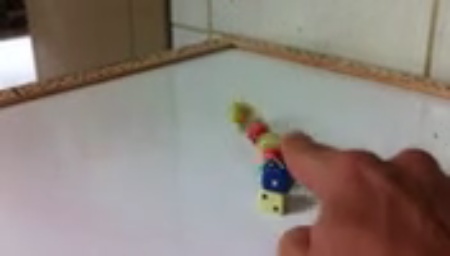}
        \includegraphics[width=0.5in]{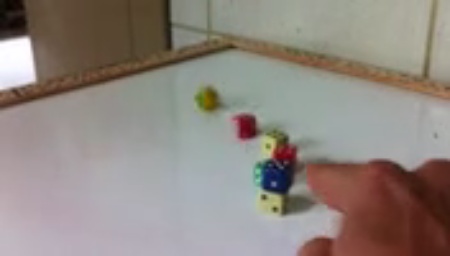}
        \includegraphics[width=0.5in]{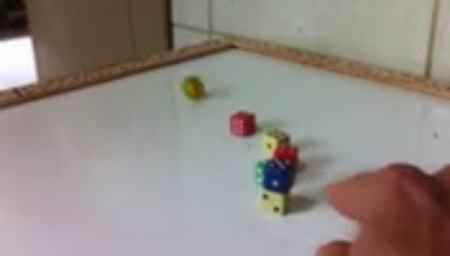}
        \includegraphics[width=0.5in]{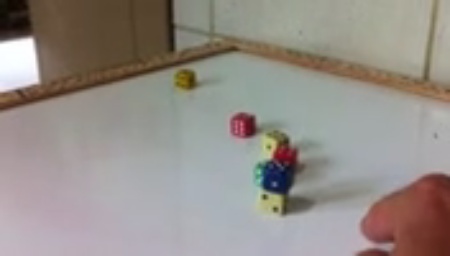}
        \includegraphics[width=0.5in]{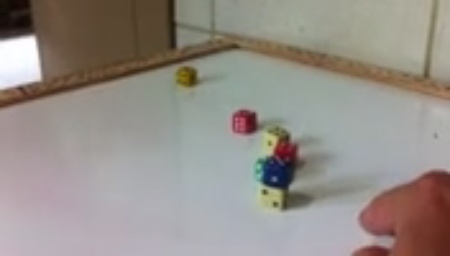}
    \end{minipage}
}\vskip -5pt
\subfigure[tearing something into two pieces]{
    \begin{minipage}[t]{1.0\linewidth}
        \centering
        \includegraphics[width=0.5in]{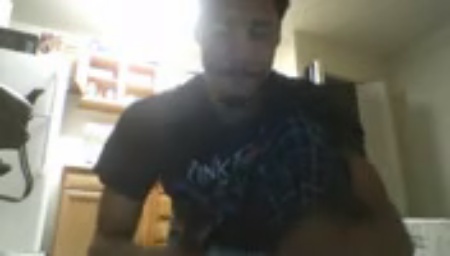}
        \includegraphics[width=0.5in]{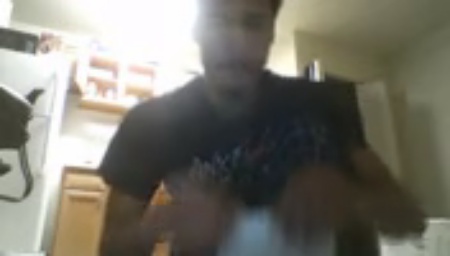}
        \includegraphics[width=0.5in]{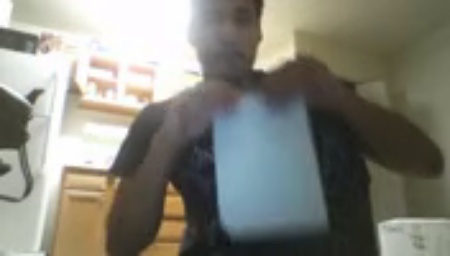}
        \includegraphics[width=0.5in]{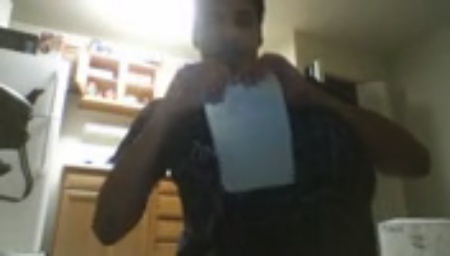}
        \includegraphics[width=0.5in]{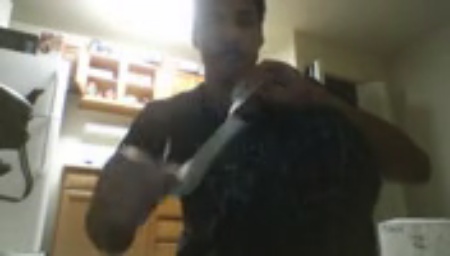}
        \includegraphics[width=0.5in]{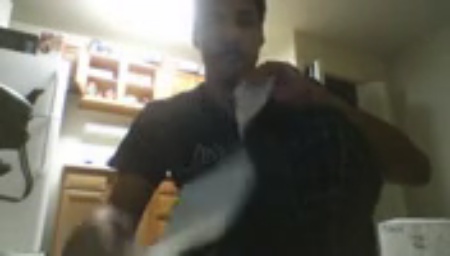}
        \includegraphics[width=0.5in]{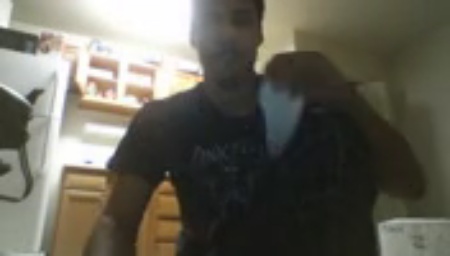}
        \includegraphics[width=0.5in]{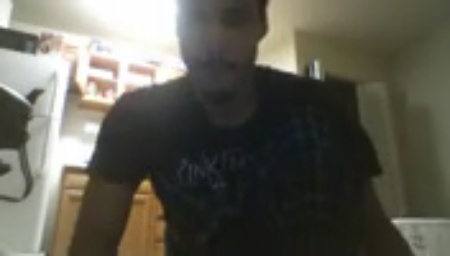}
    \end{minipage}
}\vskip -5pt
\subfigure[throwing something in the air and letting it fall]{
    \begin{minipage}[t]{1.0\linewidth}
        \centering
        \includegraphics[width=0.5in]{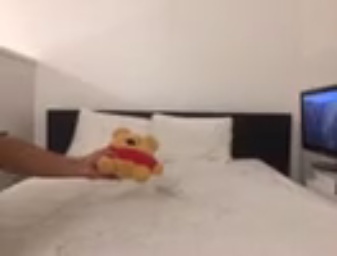}
        \includegraphics[width=0.5in]{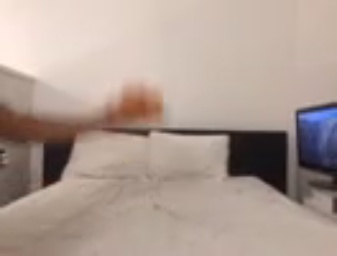}
        \includegraphics[width=0.5in]{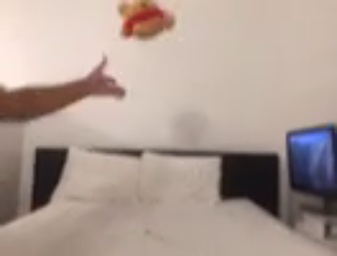}
        \includegraphics[width=0.5in]{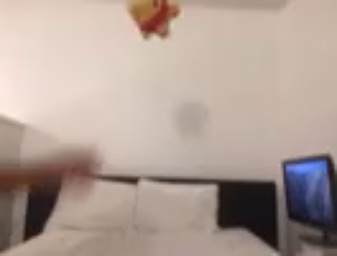}
        \includegraphics[width=0.5in]{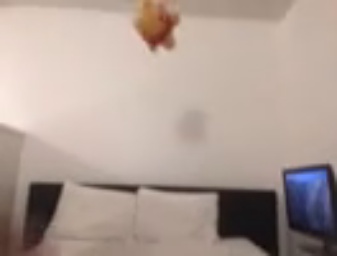}
        \includegraphics[width=0.5in]{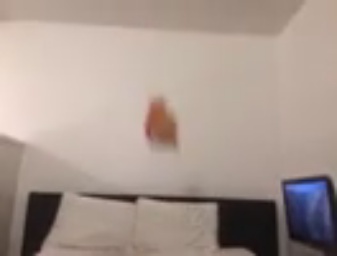}
        \includegraphics[width=0.5in]{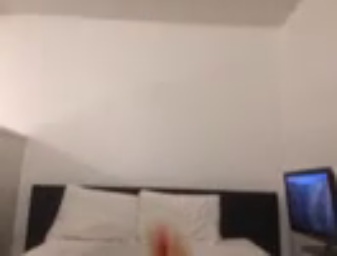}
        \includegraphics[width=0.5in]{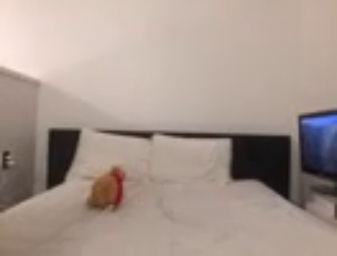}
    \end{minipage}
}
\centering
\caption{Temporal-related categories on Something-Something V1.}
\label{figure_temporal_related}
\end{figure*}

\vspace{-3pt}
\noindent\textbf{Attention for Different Datasets.} Something-Something V1 dataset is more temporal-related compared to Kinetics-400 dataset. Our network learns to focus more on temporal clues for Something-Something V1 while focus more on spatial clues for Kinetics-400, as shown in Figure \ref{figure_dataset}. We also randomly show several videos from different categories for the two datasets with high prediction accuracy, as in Figure \ref{figure_appearance_related} and Figure \ref{figure_temporal_related}. We notice that indeed Kinetics-400 dataset has more appearance-related categories (Figure \ref{figure_appearance_related}, including \texttt{\small{brushing teeth}}, \texttt{\small{carrying baby}}, etc.) while Something-Something V1 has more temporal-related action categories (Figure \ref{figure_temporal_related}, including \texttt{\small{opening something}}, \texttt{\small{tearing something into two pieces}}, etc.).

\vspace{-3pt}
\noindent\textbf{Attention for Different Modalities.} We are also interested in the network behaviour for different input modalities. We plot the attention for RGB and Flow modalities in Figure \ref{figure_modality}. It can be seen that the network pays more ``attention'' to the temporal components for Flow input, compared to RGB input. This coincides with the consensus that optical flow is more sensitive to temporal clues compared to raw RGB frames. This also partly explains the superior performance achieved by the commonly-adopted two-stream architecture.
\vspace{-30pt}

\clearpage

\clearpage

\end{document}